\ifdefined\pdfobjcompresslevel\pdfobjcompresslevel=0\relax\fi
\ifdefined\pdfsuppressptexinfo\pdfsuppressptexinfo=15\relax\fi
\ifdefined\pdfinfoomitdate\pdfinfoomitdate=1\relax\fi
\documentclass[10pt,twocolumn,letterpaper]{article}
\usepackage[pagenumbers]{cvpr}

\usepackage{cuted}

\usepackage{microtype}

\newcommand{\method}{SkillDiff\xspace}
\newcommand{\modelLuna}{GPT-5.6-Luna\xspace}
\newcommand{\modelSol}{GPT-6.1-Sol\xspace}
\newcommand{\modelMimo}{MiMo-V2.5-Pro\xspace}
\newcommand{\modelQwen}{Qwen3.7-Plus\xspace}

\newcommand{\modelQwenFlash}{Qwen3.8-Flash\xspace}
\newcommand{\modelHaiku}{Claude Haiku 5.5\xspace}
\newcommand{\modelJudge}{\modelLuna}

\usepackage{pgfplots}
\usepgfplotslibrary{groupplots}
\pgfplotsset{compat=1.18}

\definecolor{cRew}{HTML}{6983A5}
\definecolor{cSome}{HTML}{A8C6E4}
\definecolor{cOrig}{HTML}{E1B290}
\colorlet{cNone}{cOrig}

\definecolor{cAll}{HTML}{B9AEA3}
\usepackage[most]{tcolorbox}
\usepackage{colortbl}

\colorlet{hlRow}{cSome!40}
\colorlet{hlCol}{cSome!18}
\colorlet{hlAuth}{cOrig!28}

\newlength{\colw}\newlength{\colwtmp}
\newcolumntype{W}{w{c}{\colw}}
\newcommand{\fitcols}[3]{\settowidth{\colwtmp}{#2}\setlength{\colwtmp}{\dimexpr(\colwtmp-2\tabcolsep*\numexpr#1-1\relax)/#1\relax}\settowidth{\colw}{#3}\ifdim\colwtmp>\colw\setlength{\colw}{\colwtmp}\fi}
\usepackage{etoc}
\usepackage{placeins}

\definecolor{cvprblue}{rgb}{0.21,0.49,0.74}
\usepackage[pagebackref,breaklinks,colorlinks,allcolors=cvprblue]{hyperref}
\hypersetup{pdfcreator={},pdfproducer={}}

\title{How Feedback-Free Self-Evolution Reshapes Agent Skills: A Systematic Study}

\newcommand{\preprintauthors}{{\large Litao Guo\textsuperscript{1*}\quad Man Chen\textsuperscript{1*}\quad Xinli Xu\textsuperscript{1}\quad Jinsong Zhou\textsuperscript{1}\quad Zixin Zhang\textsuperscript{1,3}\quad Ying-Cong Chen\textsuperscript{1,2\textdagger}}\par\smallskip
\textsuperscript{1}The Hong Kong University of Science and Technology (Guangzhou)\par
\textsuperscript{2}The Hong Kong University of Science and Technology\par
\textsuperscript{3}Knowin\par\smallskip
{\small * Equal contribution.\qquad \textdagger\ Corresponding author.}}

\definecolor{rfblue}{HTML}{1559A8}
\definecolor{rfbluebg}{HTML}{F2F7FF}
\renewenvironment{abstract}{\begin{tcolorbox}[enhanced,colback=rfbluebg,colframe=rfblue,boxrule=.6pt,arc=2.5mm,left=7mm,right=7mm,top=3.5mm,bottom=4mm]
\centerline{\large\scshape\color{rfblue}Abstract}\vspace{1ex}\noindent\ignorespaces}{\end{tcolorbox}}
\begin{document}
\etocdepthtag.toc{mtmain}
\twocolumn[{\begin{minipage}{\textwidth}
\noindent\includegraphics[height=12mm]{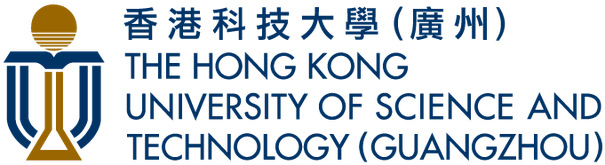}\hfill 2026-11-19\par\vspace{.09in}
{\color{rfblue}\hrule height .6pt}\vspace{.2in}
{\centering\LARGE\scshape\thetitle\par}\vspace{.18in}
{\color{rfblue}\hrule height .6pt}\vspace{.14in}
{\centering\preprintauthors\par}
\begin{abstract}
Agent skills are documents in which experts write down procedural knowledge, and models increasingly maintain them by rewriting them without feedback.
Such rewriting could adapt a skill to the model that uses it or erode the expert knowledge that makes it useful, yet how skills evolve under this feedback-free self-evolution has not been studied.
We let models rewrite expert skills round after round without any task, run or judgment and run agents built on the same models with the rewritten skills.
Our experiments show that the skills lose value rather than improve, as every model we test performs worse with them on every benchmark, by 15 points on average on SkillsBench after thirty rounds.
Sampling each model's prior with skills it writes from the topic alone, we find that rewriting pulls skills toward this prior and gives up first the knowledge outside it.
Putting this knowledge back recovers much of the lost pass rate, so an expert's skill is valuable largely for what the model does not write on its own.
Building on this insight, \method marks this knowledge in place and raises the pass rate on SkillsBench by 4.8 points on average without feedback or any change to the skill's text.
Project repository: \url{https://github.com/LitaoGuo/How-Self-Evolution-Reshapes-Skills}.
\end{abstract}
\par\medskip\noindent\includegraphics[width=\linewidth]{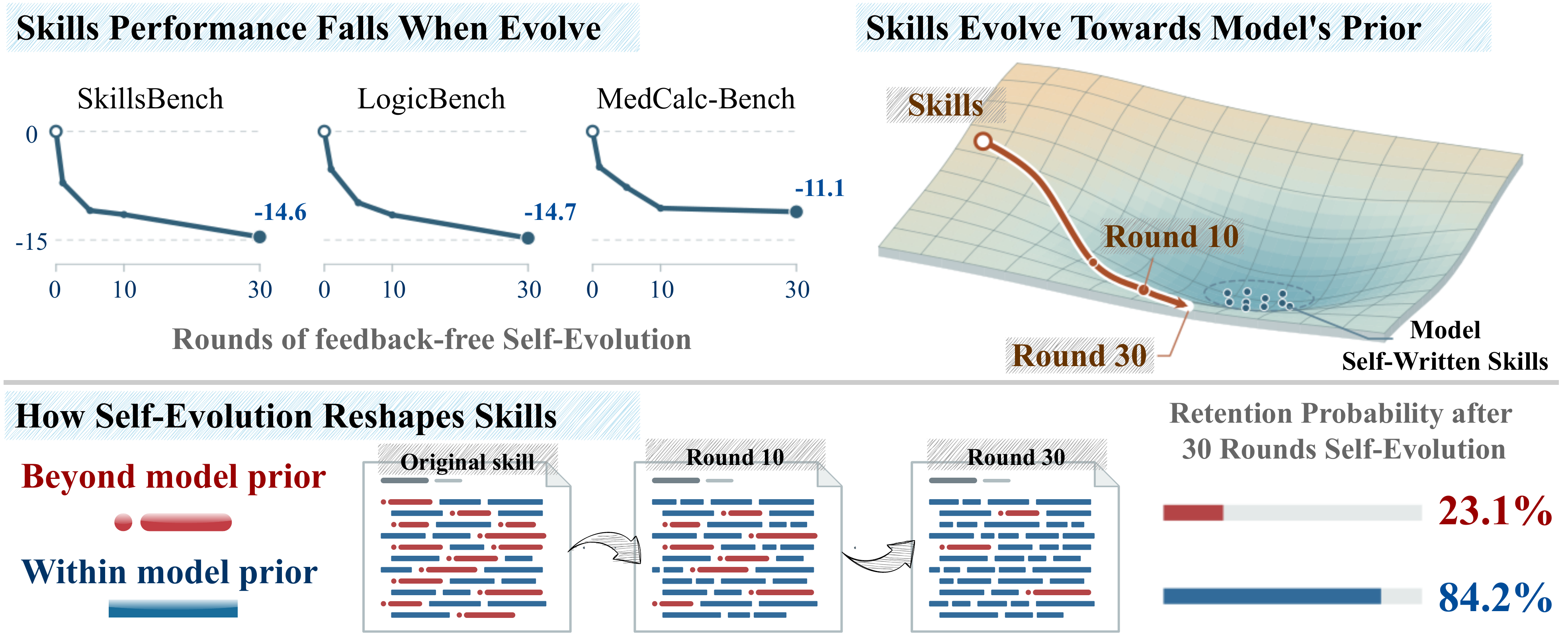}
\end{minipage}\vspace{1em}}]

\section{Introduction}
\label{sec:intro}

Large language model agents now carry out long, multi-step tasks in real software environments, from fixing code and analyzing data to working with office documents.
Such tasks depend on procedural knowledge, such as which steps to take, which tools and scripts to call and which mistakes to avoid, and a general model does not always have this knowledge for a given domain.
Skills supply it.
A skill is a short document, usually written by an expert, that an agent loads when a task calls for it, often together with scripts and reference files~\cite{anthropic_2025_agentskills}.
Within a year of the format's introduction, public repositories held millions of skill files~\cite{destefanis_2026_gitskills}, and on SkillsBench curated skills raise the average pass rate of eighteen agent configurations by 16.6 points~\cite{li_2026_skillsbench}.

Keeping this gain, however, takes more than writing a skill once.
A skill goes stale when the tools and conventions it describes change~\cite{duan_2026_repo2skillevo}.
It has to be adapted when it moves to another project~\cite{wu_2026_skilladaptation}, task or execution environment~\cite{li_2026_skillaligner}, agent framework~\cite{ouyang_2026_skcc} or model, since a skill that helps one model can hurt another~\cite{yu_2026_masa}.
Even a skill written by an expert can be incomplete or misleading for the task at hand~\cite{gao_2026_skillaudit} and lower the pass rate on some tasks~\cite{li_2026_skillsbench}, and long skills are shortened to save context~\cite{xing_2026_skillremember,gao_2026_skillreducer}.
Each of these cases calls for rewriting the skill, and recent methods leave this rewriting to an agent or a language model~\cite{duan_2026_repo2skillevo,yu_2026_masa,gao_2026_skillreducer}.

Letting agents improve their own skills has thus become an active topic in research on self-evolving agents~\cite{gao_2025_selfevolving}.
Many methods keep a rewrite only when feedback from running the agent supports it, such as task verifiers, validation tasks or execution results~\cite{shen_2026_skillfoundry,yang_2026_skillopt,wang_2026_metaskillevolve,yu_2026_masa}.
In the cases above, however, such feedback is often unavailable.
When a skill is adapted, corrected or shortened, there is often no labeled task or reliable evaluator to judge the result~\cite{gao_2026_skillaudit,peng_2026_sso}, and verifiers confine these methods to tasks that can be checked automatically~\cite{chen_2026_judgegates}.
Research and practice therefore turn to feedback-free self-evolution, improving skills without labels, external verifiers or task outcomes~\cite{peng_2026_sso,gautam_2026_skillaxe}, or letting a model rewrite a skill without testing the result~\cite{li_2026_skillaligner}.

Yet how a skill evolves when a model rewrites it without feedback has not been studied systematically.
Two outcomes are plausible.
Rewriting could improve a skill, as the model fills in what the author left out, corrects mistakes and restates the skill in a form that suits the model, much as self-refinement improves many model outputs without external feedback~\cite{madaan_2023_selfrefine} and skills adapted to a model work better for it~\cite{yu_2026_masa}.
Rewriting could also erode the expert knowledge that makes the skill useful, since models often fail to correct their own reasoning without external feedback~\cite{huang_2024_selfcorrect}, skills that agents write for themselves do not reproduce the gain of curated skills and even fall below using no skill~\cite{li_2026_skillsbench}, and shortened skills can lose the sparse operational details that agents rely on~\cite{xing_2026_skillremember}.
To find out which happens and why, we study how skills evolve as a model rewrites them round after round with no task, run or judge, so that the evolution reflects the rewriting itself, and we run agents with the rewritten skills.

On SkillsBench, we ask models to identify problems and improve each expert skill over thirty successive rewrites.
We evaluate agents built on the same models with the rewritten skills.
Our experiments show that rewriting reduces the skills' value.
After thirty rounds, every agent's pass rate falls below that of the original skill, by 15 percentage points on average.
The decline recurs on average on two subsets of SRA-Bench~\cite{su_2026_sra}.
Our evaluations show loss after the first rewrite and further decline over subsequent rounds.

To find what change in the skills drives this decline, we trace the direction in which they evolve.
Without feedback, a rewrite can draw only on the skill and on what the model brings to its topic by itself, which we call the model's prior for the skill.
To measure this prior, we ask each model to write ten skills from a one-sentence topic without seeing the expert's document.
We call these self-written skills.
Comparing complete texts across all three benchmarks, we find that every model's rewritten skills move toward its self-written skills over the first few rounds, after which the distance levels off.
Rewriting thus adapts skills to the model by pulling them toward its prior.

Similarity over the whole text, however, cannot show which of the expert's content gives way, so we follow each skill at the level of its items.
We split each original skill into items, such as list entries, paragraphs and code blocks, and count how many of the ten self-written skills state each item.
This count measures how close the item lies to the model's prior.
An item that no self-written skill states lies outside the prior, since only the author supplied it.
If rewriting pulls skills toward the prior, items outside it should be lost first.
We test this prediction by tracking item retention across thirty rewriting rounds on all three benchmarks.
We find that every model on every benchmark retains items stated by all ten self-written skills more often than items stated by none.
Averaged across models and benchmarks, after thirty rounds only 23.1\% of items outside the prior survive, while 84.2\% of items that all ten self-written skills state remain.
Rewriting thus preferentially loses content the model rarely writes on its own while retaining content it writes consistently.

Our analyses show performance declining as skills lose content outside the model's prior.
To further test whether this association reflects a causal relationship, we restore original skill items lost during rewriting and measure task performance.
Starting with skills after thirty successive rewrites, we remove conflicting versions of lost items to form a common base.
We restore the original items, producing one version with items outside the prior and another with the other lost items.
We evaluate agents with both versions on all three benchmarks.
On SkillsBench, restoring the items outside the prior recovers most of the lost pass rate, about twice as much as restoring the other items, and restoring all lost items recovers about as much as restoring these alone, although it puts back more text.
Losing the knowledge outside the prior therefore causes much of the decline.
The value of an expert's skill lies largely in this knowledge, which is exactly what rewriting without feedback gives up first.

This insight leads to a simple way of improving a skill without feedback.
Since the skill is the only source of the knowledge outside the model's prior, a rewrite can at best keep it, and what remains is to make sure the agent attends to it when it follows the skill.
We propose \method, which compares the original skill with the model's self-written skills, marks in place every item outside the prior, and adds a short note asking the agent to carry the marked items into its plan as written.
\method needs no feedback, no run of the agent and no change to a word of the skill, yet on SkillsBench it raises the pass rate by 4.8 points on average over the original skill.
That the agent gains although the skill gains no new content supports our account from the side of use: an expert's skill carries knowledge outside the model's prior, and the agent solves more tasks when this knowledge is both kept in the skill and attended to when the agent follows it.

Our contributions are as follows.
\begin{itemize}
  \item \textbf{Degradation.} We conduct the first systematic study of how agent skills evolve under feedback-free self-evolution and show that it degrades expert skills round after round instead of improving them.
  \item \textbf{Diagnosis.} We sample the model's prior for each skill as skills it writes from the topic alone and show that rewriting pulls skills toward this prior and drops first the knowledge outside it.
  \item \textbf{Cause.} We establish a causal link between the performance decline and the shift of skill content toward the model's prior.
  \item \textbf{Method.} We propose \method, a simple method that improves skills without feedback or rewriting by pointing the agent to the knowledge outside the model's prior.
\end{itemize}

\section{Feedback-Free Self-Evolution's Effect on Agent Performance}
\label{sec:effect}

\begin{figure*}[t]
\centering
\scalebox{1.00}{\begin{tikzpicture}
\begin{groupplot}[group style={group size=6 by 3, horizontal sep=0.65cm, vertical sep=0.3cm}, scale only axis, width=2.05cm, height=1.45cm, axis lines=box, axis line style={black, line width=0.5pt}, tick style={black, line width=0.4pt}, major tick length=1.6pt, tick align=outside, xtick pos=bottom, ytick pos=left, xmajorgrids, ymajorgrids, major grid style={black!40, dashed, line width=0.25pt}, tick label style={font=\scriptsize}, label style={font=\scriptsize}, title style={font=\footnotesize, yshift=-3pt, text height=1.6ex, text depth=0.4ex}]
\nextgroupplot[xmin=-2, xmax=32, xtick={0,10,20,30}, ymin=16.8, ymax=52.8, ytick={20,30,40,50}, title={\modelLuna}, ylabel={{\footnotesize SkillsBench}\\{\scriptsize Pass rate, \%}}, ylabel style={align=center}, xticklabels={}, legend to name=legdecline, legend columns=-1, legend cell align=left, legend style={font=\scriptsize, draw=none, /tikz/every even column/.append style={column sep=10pt}}]
\addplot[cOrig, line width=1.4pt, domain=-2:32, samples=2] {44.8};
\addplot[cRew, line width=1.4pt, mark=triangle*, mark size=2.2pt, mark options={solid, fill=cRew, draw=white, line width=0.25pt}] coordinates {(0,44.8) (1,33.3) (5,26.4) (10,29.9) (30,21.8)};
\addlegendentry{Original skill}\addlegendentry{Rewritten skill}
\nextgroupplot[xmin=-2, xmax=32, xtick={0,10,20,30}, ymin=27.2, ymax=63.2, ytick={30,40,50,60}, title={\modelSol}, xticklabels={}]
\addplot[cOrig, line width=1.4pt, domain=-2:32, samples=2] {55.2};
\addplot[cRew, line width=1.4pt, mark=triangle*, mark size=2.2pt, mark options={solid, fill=cRew, draw=white, line width=0.25pt}] coordinates {(0,55.2) (1,47.1) (5,46.0) (10,41.4) (30,42.5)};
\nextgroupplot[xmin=-2, xmax=32, xtick={0,10,20,30}, ymin=-1.6, ymax=34.4, ytick={0,10,20,30}, title={\modelMimo}, xticklabels={}]
\addplot[cOrig, line width=1.4pt, domain=-2:32, samples=2] {26.4};
\addplot[cRew, line width=1.4pt, mark=triangle*, mark size=2.2pt, mark options={solid, fill=cRew, draw=white, line width=0.25pt}] coordinates {(0,26.4) (1,28.7) (5,18.4) (10,25.3) (30,14.9)};
\nextgroupplot[xmin=-2, xmax=32, xtick={0,10,20,30}, ymin=11.1, ymax=47.1, ytick={20,30,40}, title={\modelQwen}, xticklabels={}]
\addplot[cOrig, line width=1.4pt, domain=-2:32, samples=2] {39.1};
\addplot[cRew, line width=1.4pt, mark=triangle*, mark size=2.2pt, mark options={solid, fill=cRew, draw=white, line width=0.25pt}] coordinates {(0,39.1) (1,29.9) (5,24.1) (10,21.8) (30,21.8)};
\nextgroupplot[xmin=-2, xmax=32, xtick={0,10,20,30}, ymin=19.1, ymax=55.1, ytick={30,40,50}, title={\modelHaiku}, xticklabels={}]
\addplot[cOrig, line width=1.4pt, domain=-2:32, samples=2] {47.1};
\addplot[cRew, line width=1.4pt, mark=triangle*, mark size=2.2pt, mark options={solid, fill=cRew, draw=white, line width=0.25pt}] coordinates {(0,47.1) (1,44.8) (5,39.1) (10,36.8) (30,34.5)};
\nextgroupplot[xmin=-2, xmax=32, xtick={0,10,20,30}, ymin=0.7, ymax=36.7, ytick={10,20,30}, title={\modelQwenFlash}, xticklabels={}]
\addplot[cOrig, line width=1.4pt, domain=-2:32, samples=2] {28.7};
\addplot[cRew, line width=1.4pt, mark=triangle*, mark size=2.2pt, mark options={solid, fill=cRew, draw=white, line width=0.25pt}] coordinates {(0,28.7) (1,14.9) (5,21.8) (10,17.2) (30,18.4)};
\nextgroupplot[xmin=-2, xmax=32, xtick={0,10,20,30}, ymin=63.6, ymax=99.6, ytick={70,80,90}, ylabel={{\footnotesize LogicBench}\\\strut}, ylabel style={align=center}, xticklabels={}]
\addplot[cOrig, line width=1.4pt, domain=-2:32, samples=2] {91.6};
\addplot[cRew, line width=1.4pt, mark=triangle*, mark size=2.2pt, mark options={solid, fill=cRew, draw=white, line width=0.25pt}] coordinates {(0,91.6) (1,82.9) (5,76.8) (10,76.7) (30,72.1)};
\nextgroupplot[xmin=-2, xmax=32, xtick={0,10,20,30}, ymin=64.1, ymax=100.1, ytick={70,80,90}, xticklabels={}]
\addplot[cOrig, line width=1.4pt, domain=-2:32, samples=2] {92.1};
\addplot[cRew, line width=1.4pt, mark=triangle*, mark size=2.2pt, mark options={solid, fill=cRew, draw=white, line width=0.25pt}] coordinates {(0,92.1) (1,81.7) (5,78.4) (10,78.0) (30,76.1)};
\nextgroupplot[xmin=-2, xmax=32, xtick={0,10,20,30}, ymin=53.8, ymax=89.8, ytick={60,70,80}, xticklabels={}]
\addplot[cOrig, line width=1.4pt, domain=-2:32, samples=2] {81.8};
\addplot[cRew, line width=1.4pt, mark=triangle*, mark size=2.2pt, mark options={solid, fill=cRew, draw=white, line width=0.25pt}] coordinates {(0,81.8) (1,81.7) (5,75.3) (10,73.2) (30,69.3)};
\nextgroupplot[xmin=-2, xmax=32, xtick={0,10,20,30}, ymin=63.8, ymax=99.8, ytick={70,80,90}, xticklabels={}]
\addplot[cOrig, line width=1.4pt, domain=-2:32, samples=2] {91.8};
\addplot[cRew, line width=1.4pt, mark=triangle*, mark size=2.2pt, mark options={solid, fill=cRew, draw=white, line width=0.25pt}] coordinates {(0,91.8) (1,87.8) (5,84.3) (10,82.5) (30,80.7)};
\nextgroupplot[xmin=-2, xmax=32, xtick={0,10,20,30}, ymin=42.8, ymax=78.8, ytick={50,60,70}, xticklabels={}]
\addplot[cOrig, line width=1.4pt, domain=-2:32, samples=2] {70.8};
\addplot[cRew, line width=1.4pt, mark=triangle*, mark size=2.2pt, mark options={solid, fill=cRew, draw=white, line width=0.25pt}] coordinates {(0,70.8) (1,68.9) (5,67.2) (10,66.1) (30,63.9)};
\nextgroupplot[xmin=-2, xmax=32, xtick={0,10,20,30}, ymin=60.8, ymax=96.8, ytick={70,80,90}, xticklabels={}]
\addplot[cOrig, line width=1.4pt, domain=-2:32, samples=2] {88.8};
\addplot[cRew, line width=1.4pt, mark=triangle*, mark size=2.2pt, mark options={solid, fill=cRew, draw=white, line width=0.25pt}] coordinates {(0,88.8) (1,82.5) (5,75.8) (10,71.2) (30,66.4)};
\nextgroupplot[xmin=-2, xmax=32, xtick={0,10,20,30}, ymin=61.1, ymax=97.1, ytick={70,80,90}, ylabel={{\footnotesize MedCalc-Bench}\\\strut}, ylabel style={align=center}, xlabel={Round}]
\addplot[cOrig, line width=1.4pt, domain=-2:32, samples=2] {89.1};
\addplot[cRew, line width=1.4pt, mark=triangle*, mark size=2.2pt, mark options={solid, fill=cRew, draw=white, line width=0.25pt}] coordinates {(0,89.1) (1,86.0) (5,83.1) (10,83.0) (30,85.5)};
\nextgroupplot[xmin=-2, xmax=32, xtick={0,10,20,30}, ymin=62.2, ymax=98.2, ytick={70,80,90}]
\addplot[cOrig, line width=1.4pt, domain=-2:32, samples=2] {90.2};
\addplot[cRew, line width=1.4pt, mark=triangle*, mark size=2.2pt, mark options={solid, fill=cRew, draw=white, line width=0.25pt}] coordinates {(0,90.2) (1,79.3) (5,75.3) (10,74.2) (30,73.7)};
\nextgroupplot[xmin=-2, xmax=32, xtick={0,10,20,30}, ymin=57.5, ymax=93.5, ytick={60,70,80,90}]
\addplot[cOrig, line width=1.4pt, domain=-2:32, samples=2] {85.5};
\addplot[cRew, line width=1.4pt, mark=triangle*, mark size=2.2pt, mark options={solid, fill=cRew, draw=white, line width=0.25pt}] coordinates {(0,85.5) (1,83.8) (5,86.0) (10,80.1) (30,79.5)};
\nextgroupplot[xmin=-2, xmax=32, xtick={0,10,20,30}, ymin=63.6, ymax=99.6, ytick={70,80,90}]
\addplot[cOrig, line width=1.4pt, domain=-2:32, samples=2] {91.6};
\addplot[cRew, line width=1.4pt, mark=triangle*, mark size=2.2pt, mark options={solid, fill=cRew, draw=white, line width=0.25pt}] coordinates {(0,91.6) (1,83.5) (5,81.5) (10,79.5) (30,81.5)};
\nextgroupplot[xmin=-2, xmax=32, xtick={0,10,20,30}, ymin=63.4, ymax=99.4, ytick={70,80,90}]
\addplot[cOrig, line width=1.4pt, domain=-2:32, samples=2] {91.4};
\addplot[cRew, line width=1.4pt, mark=triangle*, mark size=2.2pt, mark options={solid, fill=cRew, draw=white, line width=0.25pt}] coordinates {(0,91.4) (1,89.1) (5,84.7) (10,81.5) (30,74.0)};
\nextgroupplot[xmin=-2, xmax=32, xtick={0,10,20,30}, ymin=62.6, ymax=98.6, ytick={70,80,90}]
\addplot[cOrig, line width=1.4pt, domain=-2:32, samples=2] {90.6};
\addplot[cRew, line width=1.4pt, mark=triangle*, mark size=2.2pt, mark options={solid, fill=cRew, draw=white, line width=0.25pt}] coordinates {(0,90.6) (1,87.2) (5,81.5) (10,76.5) (30,77.6)};
\end{groupplot}
\node[anchor=south, yshift=1pt] at (current bounding box.north) {\pgfplotslegendfromname{legdecline}};
\end{tikzpicture}}
\caption{\textbf{By round 30, rewriting lowers every agent's SkillsBench pass rate.} Original and rewritten skills, with the same model used for rewriting and evaluation.}
\label{fig:decline}
\end{figure*}
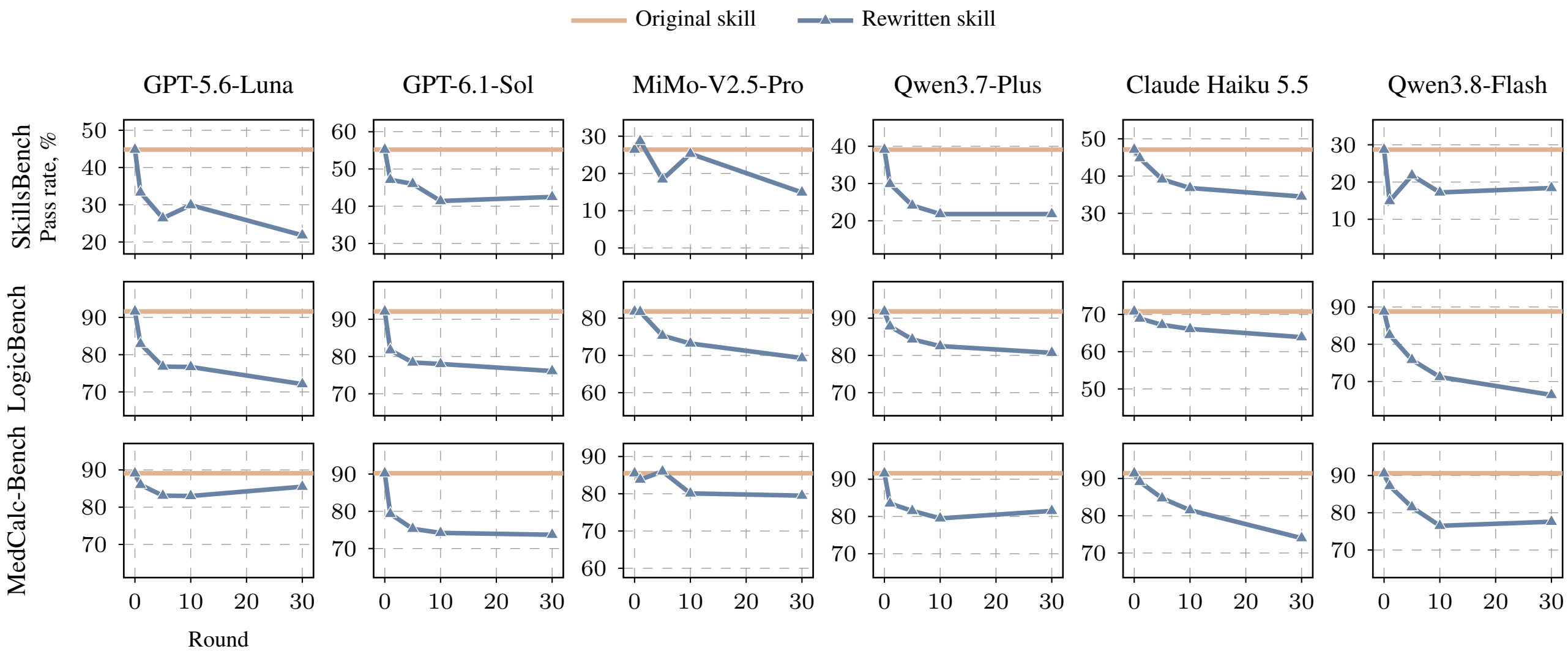

A model may improve an expert skill by filling gaps and adapting its instructions, or erode the knowledge that makes it useful.
We distinguish these outcomes by evaluating agents with skills that models recursively rewrite without feedback.

\subsection{Study Design}
\label{sec:design}

\noindent\textbf{Feedback-free self-evolution.}\enspace
Each model rewrites a skill from its text and a one-sentence topic, with an instruction to identify problems and improve the skill.
We provide no task, outcome or judgment.
Each rewrite replaces its predecessor for thirty recursive rounds to trace where repeated rewriting leads.
The same model then uses its rewritten skills as an agent, giving rewriting an opportunity to adapt the skill to that model.

\noindent\textbf{Benchmarks.}\enspace
We choose benchmarks that pair their tasks with expert-written skills, check the outcome of every task automatically, and differ in how the model reaches the skill.
SkillsBench~\cite{li_2026_skillsbench} contains 87 tasks across eight domains, from natural science and finance to office work and software engineering.
Each task runs in its own container with fixed data and a deterministic verifier, and the agent works through many steps, loading a skill when it judges the skill relevant.
Most tasks, 63 of the 87, come with several curated skills, and choosing among them is a problem separate from what a skill contains.
We set this choice aside: for each of these tasks, GPT-5.6-Sol~\cite{openai_2026_gpt56} picks the most relevant skill from the task instruction and the skills alone, before any run, and the agent receives that skill, so that each task tests one expert skill.
Only the skill document is rewritten, and the scripts and reference files that come with the skill stay unchanged.
LogicBench~\cite{parmar_2024_logicbench} and MedCalc-Bench~\cite{khandekar_2024_medcalc} are two subsets of SRA-Bench~\cite{su_2026_sra}, which pairs its questions with manually constructed gold skills.
LogicBench holds 760 logical reasoning questions with 19 skills, and MedCalc-Bench 1,100 questions that ask for a medical calculation from a patient note, with 55 skills.
The model answers each question in a single call with its gold skill in the prompt, so the skill takes part in one answer rather than in a plan of many steps.

\noindent\textbf{Models and harnesses.}\enspace
We choose models from several families and of different strengths, so that the result does not hinge on one model, \modelLuna~\cite{openai_2026_gpt56} and \modelSol~\cite{openai_2026_gpt61sol}, \modelMimo~\cite{mimo2026v25pro}, \modelHaiku~\cite{anthropic_2026_haiku55}, and \modelQwen~\cite{qwen37plus} together with \modelQwenFlash~\cite{qwen3.8flashnext}.
With the original skills, their pass rates on SkillsBench range from about a quarter to more than half of the tasks.
Each model runs as an agent in a harness that loads skills on demand, as skills are used in practice, the GPT models in Codex, \modelHaiku in Claude Code, and the other models in OpenHands~\cite{wang_2024_openhands}.

\noindent\textbf{Evaluation protocol.}\enspace
Every agent runs each SkillsBench task in its container with the skill placed where the harness loads skills, and a task passes when its verifier accepts the final state of the environment.
For every model, we evaluate the original skill and the skills of rounds 1, 5, 10 and 30 on the same tasks.
On LogicBench and MedCalc-Bench, we also let every model answer without any skill, and accuracy is the share of questions answered correctly.

\subsection{Rewriting Lowers the Pass Rate}
\label{sec:effect_results}

Following this setup, we evaluate all six models on the three benchmarks using the original skills and skills from different rewriting rounds.

\noindent\textbf{Every agent loses pass rate, most of all in the first rounds.}\enspace
Even the first rewrite lowers the SkillsBench pass rate by 7.1 points on average across agents (\cref{fig:decline}).
The loss grows to 10.9 points after five rounds and 14.6 after thirty.
By round thirty, every agent performs worse than with the original skill, by 10.3 to 23.0 points.
Rewriting thus wears the skill down from the first round, fastest in the early rounds and more slowly afterwards.

\noindent\textbf{After thirty rounds, the skill is worth about as much as having none.}\enspace
On LogicBench and MedCalc-Bench, where the model reads the whole skill once in its prompt, thirty rounds of rewriting leave almost none of the original skill's gain.
On LogicBench, mean accuracy falls from 86.2\% with the original skill to 71.4\%, matching the level without a skill.
On MedCalc-Bench, it falls from 89.7\% to 78.7\%, below the 81.1\% achieved without a skill.
Accuracy falls for every model on both benchmarks, by 14.7 and 11.1 points on average.

\noindent\textbf{The model's own rewrite serves the model worse.}\enspace
Feedback-free rewriting reduces the value of expert skills both during multi-step use and in a single prompt.
This decline persists when the same model rewrites and uses the skill, showing that adaptation by its own judgment serves it worse than the expert's original.

\section{Deep Analysis}
\label{sec:analysis}

We trace the decline at two levels: the direction of change over the whole skill and the content that individual items lose.

\subsection{Rewritten Skills Move Toward the Model's Prior}
\label{sec:direction}

\input{fig/distance}

Without feedback, a rewrite can draw only on the skill itself and on what the model already holds about the skill's topic, the model's prior for the skill.
Declining performance shows that rewriting reduces a skill's value, but leaves the direction of textual change unclear.
We therefore compare rewritten skills with skills the same model writes from the topic alone, testing whether rewriting moves toward the model's own content.

\noindent\textbf{Sampling the prior.}\enspace
For each skill, we sample the model's prior with ten self-written skills.
The model sees the same one-sentence topic and length target as in rewriting, without the expert's document.
Matching the topic and length lets these samples represent what the model supplies on its own.

\noindent\textbf{Measuring the distance.}\enspace
We use TF-IDF word vectors because they give greater weight to distinctive terms, such as the expert's commands, paths and parameter names.
A skill's distance is its mean cosine distance to the self-written skills, divided by their mean pairwise distance.
At 1, a skill is as close to the self-written skills as they are to one another.

\noindent\textbf{Rewriting pulls the skills toward the prior.}\enspace
We measure these distances for the original skills and each model's rewrites across all three benchmarks, tracking their change over rewriting rounds.
As shown in \cref{fig:distance}, the original skills lie 1.2 to 1.7 units from the self-written skills, and rewriting brings the skills of every model closer to them on every benchmark.
The movement is gradual but comes to an end.
On average over models and benchmarks, the distance falls from 1.37 by ever smaller steps over the first rounds, reaches 1.19 by round five and changes little afterwards, standing at 1.22 after thirty rounds.
The rewriting thus adapts the skill to the model, as the first outcome expects, but by pulling the skill toward what the model would write on its own, until the skill settles between the original and the self-written skills.

\noindent\textbf{The distance alone does not predict the loss.}\enspace
After the fifth round the distance changes little, yet the pass rate keeps falling, so the distance over the whole text does not track what the later rounds still remove.
A skill can move far toward the prior while keeping the few items that its questions depend on, or move little and lose one of them, so the distance over the whole text cannot show which of the expert's content gives way.

\subsection{Content Outside the Prior Is Lost First}
\label{sec:selection}

\input{fig/coverage}
\begin{table}[t]
\centering\footnotesize
\setlength{\tabcolsep}{2pt}
\resizebox{\columnwidth}{!}{\fitcols{3}{MedCalc-Bench}{$+74$}
\begin{tabular}{@{}l>{\columncolor{hlAuth}}WW>{\columncolor{hlCol}}W>{\columncolor{hlAuth}}WW>{\columncolor{hlCol}}W>{\columncolor{hlAuth}}WW>{\columncolor{hlCol}}W@{}}
\toprule
& \multicolumn{3}{c}{SkillsBench} & \multicolumn{3}{c}{LogicBench} & \multicolumn{3}{c}{MedCalc-Bench}\\
\cmidrule(lr){2-4}\cmidrule(lr){5-7}\cmidrule(l){8-10}
Model & 0/10 & 10/10 & $\Delta$ & 0/10 & 10/10 & $\Delta$ & 0/10 & 10/10 & $\Delta$\\
\midrule
\modelLuna & 10 & 83 & $+73$ & 19 & 80 & $+61$ & 20 & 85 & $+65$\\
\modelSol & 7 & 81 & $+74$ & 11 & 82 & $+71$ & 15 & 85 & $+70$\\
\modelMimo & 32 & 90 & $+58$ & 27 & 90 & $+63$ & 21 & 90 & $+69$\\
\modelQwen & 21 & 86 & $+65$ & 22 & 88 & $+66$ & 38 & 87 & $+49$\\
\modelHaiku & 37 & 82 & $+45$ & 56 & 92 & $+36$ & 52 & 88 & $+36$\\
\modelQwenFlash & 6 & 75 & $+69$ & 16 & 74 & $+58$ & 5 & 76 & $+71$\\
\midrule
\rowcolor{hlRow}Average & 19 & 83 & $+64$ & 25 & 84 & $+59$ & 25 & 85 & $+60$\\
\bottomrule
\end{tabular}}
\caption{\textbf{After thirty rounds, 10/10 items survive more often than 0/10 items.} Share of original skill items retained after thirty rounds, in percent. The labels 0/10 and 10/10 mean that none or all ten self-written skills state an item. $\Delta$ is the difference in retention between the 10/10 and 0/10 groups, in percentage points. The final row averages retention across models.}
\label{tab:loss}
\end{table}

The distance over the whole text shows where the rewriting goes but not what it gives up, since it levels off after a few rounds while the pass rate keeps falling.
What an agent gains from a skill, however, lies in specific pieces of content, a step, a command, a rule or a fact.
We therefore move from the whole text to these pieces and ask two questions that the distance cannot answer, which of the expert's content survives the rewriting, and whether what is lost is what the model would not have written by itself.

\noindent\textbf{Splitting a skill into items.}\enspace
We split each skill mechanically into headings, list entries, code blocks, tables, paragraphs and front matter.
Every non-blank line belongs to exactly one item, covering the whole skill without overlap.
This gives 6,605 items for SkillsBench and 2,376 for LogicBench and MedCalc-Bench together.

\noindent\textbf{Judging what a document says about an item.}\enspace
Whether an item survives cannot be read from the strings, since a rewrite rephrases much of what it keeps, and it may keep names and numbers while changing what it says about them.
An LLM judge, \modelJudge for every model, therefore reads all items of an original skill together with one other document and labels every item in a single call.
For each original item, we judge whether self-written skills state the same content and whether rewrites retain it in full.
Both judgments compare meaning.
Shared words alone do not count as a match.
A rewrite keeps an item when it preserves its full meaning, and loses it when the content is absent, incomplete or changed.
The judge records content retention without evaluating whether an edit is beneficial.
We exclude items the judge cannot decide.
The same judge and the same instructions serve every model, so the comparison between items does not depend on which model rewrote them.

\noindent\textbf{Measuring how close an item lies to the prior.}\enspace
Each self-written skill is one sample of the prior, so the number of the ten self-written skills that state an item, from zero to ten, measures how readily the model writes the item on its own.
We label original items absent from all ten self-written skills as outside the prior, identifying content the model rarely writes on its own for the given topic.
Items stated in all ten represent content the model writes more consistently on its own.
The two questions now become testable.
If rewriting merely corrected and completed a skill, whether an item survives would not depend on this count, whereas if rewriting pulls the skill toward the prior, the items farthest from it should be lost first.

\noindent\textbf{Survival grows with closeness to the prior.}\enspace
We group original skill items by how many of the ten self-written skills state them and compare retention after rounds one and thirty.
As shown in \cref{fig:coverage}, on every benchmark every model keeps the items that all ten self-written skills state far more often than the items that none states, and on SkillsBench the share kept rises from each group to the next for every model.
Thirty rounds of rewriting create a sharp divide, losing most items outside the prior while retaining most items stated in all ten self-written skills.
Averaged across the six models and three benchmarks, only 23.1\% of the former survive, compared with 84.2\% of the latter, a mean retention gap of 61.1 percentage points.
The split appears from the first round on and within list entries, paragraphs, code blocks and headings alike, so it does not reflect which kinds of content a rewrite tends to shorten.
To check whether the result depends on the judge, we ask \modelSol to repeat the item judgments on 20 SkillsBench tasks, using \modelLuna's self-written and rewritten skills and the same instructions.
Both judges find much greater loss outside the prior than among the other items, with gaps of 49.5 and 52.5 percentage points under the original and second judge, respectively.
The selective loss thus persists under a different judge.
The drift toward the prior is thus selective, as rewriting keeps what the model could have supplied itself and gives up first what only the expert supplied.
\subsection{Restoring This Content Recovers the Pass Rate}
\label{sec:cause}

\input{fig/restore}
\begin{table}[t]
\centering\footnotesize
\setlength{\tabcolsep}{3pt}
\fitcols{3}{Round 30, lost items restored}{80.6}
\begin{tabular}{@{}lccW>{\columncolor{hlAuth}}WW@{}}
\toprule
& & & \multicolumn{3}{c}{Round 30, lost items restored}\\
\cmidrule(l){4-6}
Benchmark & Original & Round 30 & $\geq$1/10 & 0/10 & All\\
\midrule
SkillsBench & 40.2 & 25.7 & 30.1 & 34.5 & 35.8\\
LogicBench & 86.2 & 71.4 & 75.0 & 80.4 & 80.6\\
MedCalc-Bench & 89.7 & 78.7 & 80.0 & 83.1 & 84.4\\
\bottomrule
\end{tabular}
\caption{\textbf{Restoring 0/10 items recovers more performance than restoring $\geq$1/10 items on every benchmark.} Mean pass rates across agents, in percent, for original skills, round-30 skills and the three restored versions. The 0/10 and $\geq$1/10 groups contain lost items stated by none or at least one of ten self-written skills; All restores every lost item.}
\label{tab:restore}
\end{table}

The content outside the prior is the content that rewriting gives up first.
We now test whether losing it causes the decline.
If it does, restoring this content to the round-30 skills should recover more of the pass rate than restoring the other lost content.

\noindent\textbf{Putting the lost items back.}\enspace
To compare the effects of restoring different groups of lost items on task performance, we start from each skill after thirty rewrites.
The rewritten skill often still carries a weakened or changed version of a lost item, and restoring the original next to it would leave two conflicting versions in the same skill.
We therefore remove the sentences that carry such versions, yielding a common base for the restored versions of each skill.
We restore lost items at corresponding positions in this base, appending them to the end when we cannot locate a position.

\noindent\textbf{Three versions of each skill.}\enspace
We build three restored versions of each skill, which share the base and differ only in which lost items return.
The first puts back only the items outside the prior, the content whose effect we test.
The second puts back only the other lost items, which lie within the prior, through the same procedure, and serves as the comparison, since if the content outside the prior matters more than the other lost content, the first version should recover more than the second.
The third puts back all lost items and with them the most text, which shows whether a recovery grows with how much text returns rather than with what returns.

\noindent\textbf{Restoring the content outside the prior recovers much of the loss.}\enspace
We run agents with all three restored versions on all three benchmarks, using the same tasks and evaluation protocol as for the original and round-30 skills (\cref{fig:restore,tab:restore}).
On SkillsBench, rewriting cost 14.6 points on average across agents.
Restoring only the items outside the prior recovers 8.8 points, twice the 4.4 points recovered by restoring the other lost items.
Even restoring all lost items recovers only 10.2 points, close to the recovery from restoring the items outside the prior alone.
On LogicBench and MedCalc-Bench, too, restoring the items outside the prior recovers more than restoring the other items on average.
Losing the knowledge outside the model's prior therefore causes much of the decline.

\noindent\textbf{Summary.}\enspace
Combining these analyses, we arrive at three observations.
First, rewriting moves every model's skills toward its prior step by step, until they settle near it.
Second, the movement is selective, as rewriting gives up first the content outside the prior and keeps the content the model would write anyway.
Third, putting this content back recovers much of the lost pass rate, so losing it causes much of the decline.
The value of an expert's skill thus lies largely in the knowledge outside the model's prior, which the model does not supply by itself and which rewriting without feedback gives up first.

\section{Improving Skills Without Rewriting}
\label{sec:method}

The value of an expert's skill lies largely in the knowledge outside the model's prior.
Without feedback, a rewrite cannot add this knowledge, since the skill is its only source, and can at best keep it.
The original skill already holds it, so what remains is whether the agent uses it.
An agent may pass over these items when it follows a skill and fall back on what it would do by itself, much as a rewrite falls back on what the model would write.
We therefore propose \method, a simple method that improves a skill without feedback and without changing its text, by finding the items outside the model's prior and pointing the agent to them.

\subsection{Marking the Items Outside the Prior}
\label{sec:marking}

\begin{figure}[t]
\centering
\includegraphics[width=\columnwidth]{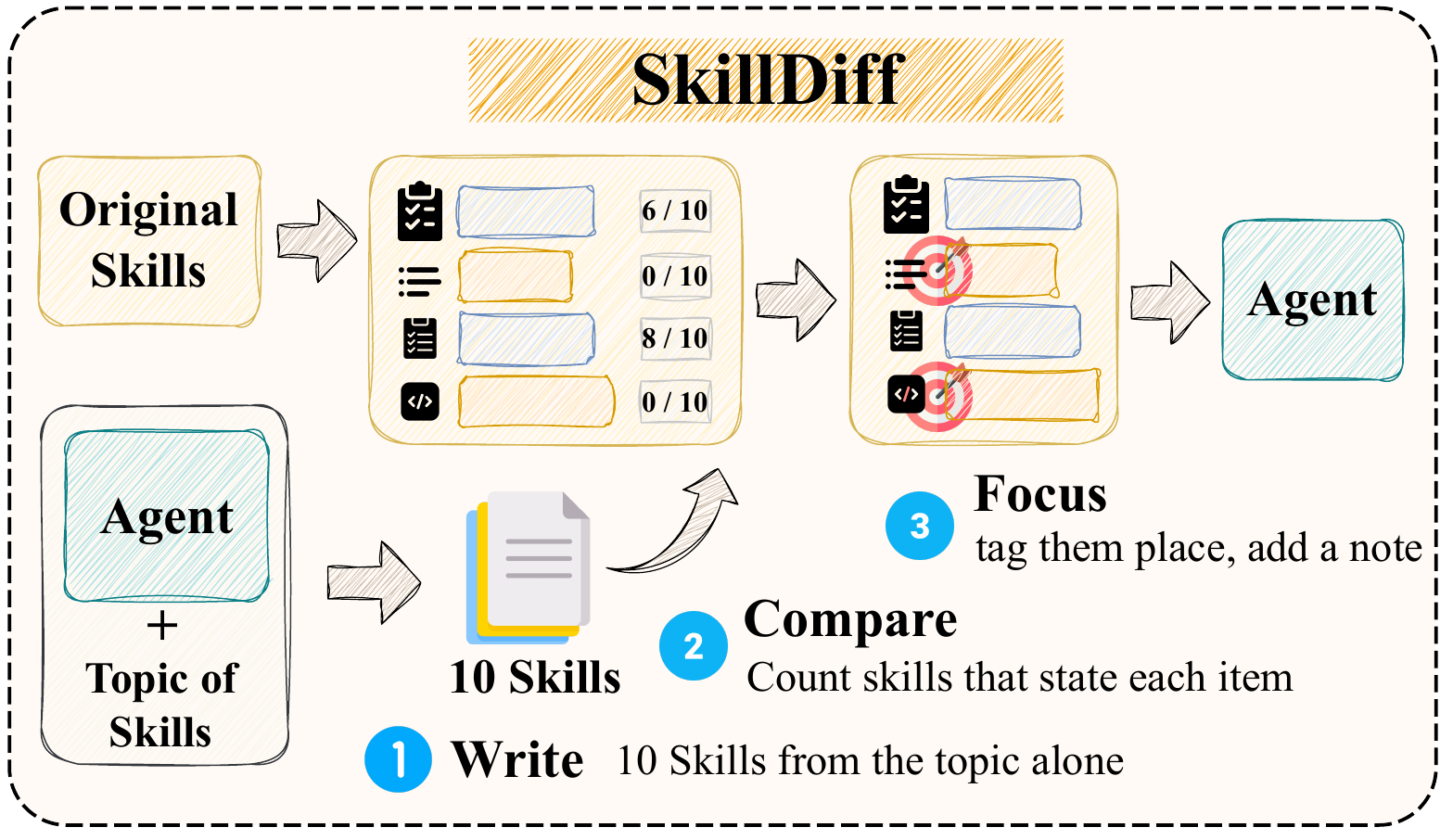}
\caption{\textbf{\method.} Generate ten skills from the topic alone, compare each original item against them, and mark in place the items that none states. A note asks the agent to carry these items into its plan as written.}
\label{fig:method}
\end{figure}

Pointing the agent to these items raises three questions, which items to point to, how to point to them without changing what the skill says, and what to ask of the agent.

\noindent\textbf{Which items.}\enspace
\method finds these items in the same way as the analysis of the rewriting.
The model writes ten self-written skills from the topic of the skill, the original skill is split into items, and the judge decides for every self-written skill whether it states each item.
\method then selects every list entry, paragraph, code block and table that none of the ten states, the items outside the prior.
Since the self-written skills come from the topic alone and the judge reads only text, the selection needs neither feedback nor a run of the agent.

\noindent\textbf{How to point.}\enspace
\method marks each selected item in place with a bold tag, \textbf{[FOCUS]}, so that every item stays in the context its author gave it, and adds a note below the title that explains the tag.
Removing the tags and the note gives back the original skill byte for byte.

\noindent\textbf{What to ask.}\enspace
The note asks the agent to pay particular attention to the marked items and to carry their specific procedures, tools, parameters, conditions and exceptions into its plan as written, instead of simplifying them to a generic approach.
The request aims at the way an agent may pass over these items when it acts, replacing the expert's specific procedure with the generic one it would choose by itself, as a rewrite replaces it with what the model would write.
\method thus needs no feedback and keeps every word of the skill in place.

\subsection{Marking Raises the Pass Rate}
\label{sec:marking_results}

\begin{table}[t]
\centering\footnotesize
\setlength{\tabcolsep}{6pt}
\begin{tabular}{@{}lc@{}}
\toprule
Model & $\Delta$ pass rate, pp\\
\midrule
\modelLuna & $+12.6$\\
\modelSol & $+5.7$\\
\modelMimo & $+3.4$\\
\modelQwen & $+2.3$\\
\modelHaiku & $+3.4$\\
\modelQwenFlash & $+1.1$\\
\midrule
\rowcolor{hlRow}Average & $+4.8$\\
\bottomrule
\end{tabular}
\caption{\textbf{Marking items outside the prior improves every agent on SkillsBench.} Pass-rate gain in percentage points over the original skill, evaluated in the same batch.}
\label{tab:marking}
\end{table}

\noindent\textbf{Evaluation.}\enspace
We evaluate \method on SkillsBench with the same agents and protocol as the rewritten skills.
\method leaves the text of the skill unchanged and only adds the tags and the note, so the comparison that matters is with the original skill itself.
The pass rate of the same skill shifts between runs made at different times, so we compare each marked skill with the original skill run in the same batch, which keeps a drift between batches out of the difference.

\noindent\textbf{Marking raises the pass rate of every agent.}\enspace
Keeping the original skill text and adding only the tags and note raises the SkillsBench pass rate by 4.8 points on average, with gains for all six agents.
This result shows that highlighting knowledge outside the model's prior and asking the agent to attend to it can improve task performance without rewriting the skill.

\section{Discussion}
\label{sec:discussion}

Our results make preserving knowledge outside the model's prior a practical concern for self-evolving agents.
This content contributes to task performance, yet rewriting tends to remove it and rarely restores it once lost.
\method shows that drawing the agent's attention to this existing knowledge can improve performance while preserving the original text.

External feedback can supply new information and guide useful changes.
Success on evaluated tasks does not guarantee that a rewrite preserves knowledge needed elsewhere.
ACE also reports loss of detail under feedback-guided rewriting and addresses it with incremental updates~\cite{zhang_2025_ace}.
These findings motivate protecting content outside the model's prior during skill updates and, when feedback is available, directing edits to the problems it identifies.

\section{Related Work}
\label{sec:related}

\paragraph{Agent skills.}
Skills provide procedural knowledge through documents and scripts that agents load on demand~\cite{anthropic_2025_agentskills}.
Agents also distill experience into code libraries~\cite{wang_2023_voyager,wang_2025_programmatic,zheng_2025_skillweaver}, workflows~\cite{wang_2024_workflow}, insights~\cite{zhao_2023_expel} and manuals~\cite{chen_2024_automanual}.
SkillsBench and SkillLearnBench compare curated skills with agent-written or continually learned skills~\cite{li_2026_skillsbench,zhong_2026_skilllearnbench}, while SRA-Bench evaluates skills retrieved into model prompts~\cite{su_2026_sra}.
Work on project adaptation~\cite{wu_2026_skilladaptation}, software maintenance~\cite{duan_2026_repo2skillevo}, model adaptation~\cite{yu_2026_masa} and compression~\cite{xing_2026_skillremember,gao_2026_skillreducer} motivates updating skills after their creation.
We study what repeated rewriting preserves or removes from expert skills and how these changes affect task performance.

\paragraph{Skill evolution.}
Self-improving agents revise prompts, memories and code using task scores or execution feedback~\cite{yang_2023_opro,agrawal_2025_gepa,shinn_2023_reflexion,zhang_2025_ace,zhang_2025_dgm,gao_2025_selfevolving}.
SkillFoundry, SkillOpt, MetaSkill-Evolve and MASA keep skill rewrites when verifiers, validation tasks or environment feedback support them~\cite{shen_2026_skillfoundry,yang_2026_skillopt,wang_2026_metaskillevolve,yu_2026_masa}.
Methods that rely on verifiers require tasks with automatically checkable outcomes~\cite{chen_2026_judgegates}.
Other methods reduce dependence on external outcome supervision while still evaluating executions.
SelfEvoSkill compares paired runs with and without the skill~\cite{gao_2026_skillaudit}; Self-Supervised Skill Optimization uses an LLM judge on unlabeled tasks~\cite{peng_2026_sso}; SkillAxe grades runs against rules drawn from the skill~\cite{gautam_2026_skillaxe}.
SkillAligner adapts retrieved skills before execution without testing the rewrite~\cite{li_2026_skillaligner}, and Hermes Agent edits skills without a test by default~\cite{nousresearch_2026_hermes}.
SkillLearnBench revises model-generated skills from the agent's executions and reports accuracy first rising and then falling in a four-round side experiment~\cite{zhong_2026_skilllearnbench}.
We isolate rewriting with no task execution or evaluation to guide the updates.
By tracking expert skills over thirty rounds, we connect performance decline to the loss of specific original content and test this connection through restoration.

\paragraph{Skill content.}
Studies of skill content explain its value through the knowledge it supplies, the procedures it encourages and the contribution of individual items.
Agent-written skills fail to reproduce the gains of curated skills~\cite{li_2026_skillsbench}, and model-generated data-science skills perform no better than the task prompt alone~\cite{huang_2026_dsskills}.
Other studies emphasize procedural guidance~\cite{jiang_2026_demystifying}, report comparable benefits from random and curated rule files for coding agents~\cite{zhang_2026_guardrails}, or link gains from revisions to newly specified commands and paths~\cite{wang_2026_skillrevisions}.
Shapley valuations estimate each skill unit's contribution through counterfactual agent runs~\cite{li_2026_skillsv,liu_2026_skillshapley}.
We examine a complementary property: how readily the model writes each item from the topic alone.
This property applies to procedures and facts alike and identifies content that rewriting preferentially loses and whose restoration recovers performance.

Iterated learning predicts convergence to a learner's prior under posterior sampling~\cite{griffiths_2007_iterated,kalish_2007_iterated}.
Language-model transmission chains drift toward model biases~\cite{acerbi_2023_transmission,perez_2024_telephone,ren_2024_biasamplification}, and recursive training on model outputs loses rare content~\cite{shumailov_2024_collapse}.
These findings motivate examining the direction of content change.
We measure that change in expert skills at the item level and use task evaluations and restoration experiments to determine its consequences for agent performance.

\section{Conclusion}
\label{sec:conclusion}

Across three benchmarks, our study of thirty rounds of feedback-free rewriting finds that task performance declines on average from the first round.
The decline holds both when agents consult skills over multiple steps and when models read them once in a prompt.
On LogicBench and MedCalc-Bench, thirty rewrites leave skills, on average, worth about as much as having none.
Rewriting moves skills toward what the model would write on its own and preferentially removes content outside its prior.
On average, restoring this content recovers more performance than restoring the other lost content, linking the selective loss to the decline.
Guided by this finding, \method marks the same content in the original skill and improves pass rates on SkillsBench without rewriting.
These results identify knowledge the model rarely writes on its own as a valuable part of expert skills and show the benefits of preserving it and directing the agent's attention to it.

\par

\par

\clearpage

\twocolumn[{\centering
{\Large\textbf{\thetitle}\\[0.5em]Supplementary Material\\[1.2em]}
\begin{minipage}{0.86\textwidth}
\etocdepthtag.toc{mtappendix}\etocsettagdepth{mtmain}{none}\etocsettagdepth{mtappendix}{subsection}\etocsettocstyle{\noindent{\large\textbf{Appendix Contents}}\par\vspace{0.4em}}{}\small\tableofcontents
\end{minipage}
\vspace{1.4em}
}]
\appendix
\raggedbottom

\section{Detailed Results}
\label{sec:suppl_results}

This section reports the analysis of the items model by model, checks the judge behind it, and examines the measure of the distance to the prior.

\subsection{Performance and Distance Endpoints}
\label{sec:suppl_endpoints}

\Cref{tab:decline,tab:distance} report the model-level endpoints of the performance and distance trajectories in \cref{fig:decline,fig:distance}.

\begin{table}[t]
\centering\footnotesize
\setlength{\tabcolsep}{3pt}
\fitcols{2}{MedCalc-Bench}{$-22.4$}
\begin{tabular}{@{}lW>{\columncolor{hlCol}}WW>{\columncolor{hlCol}}WW>{\columncolor{hlCol}}W@{}}
\toprule
& \multicolumn{2}{c}{SkillsBench} & \multicolumn{2}{c}{LogicBench} & \multicolumn{2}{c}{MedCalc-Bench}\\
\cmidrule(lr){2-3}\cmidrule(lr){4-5}\cmidrule(l){6-7}
Model & R1 & R30 & R1 & R30 & R1 & R30\\
\midrule
\modelLuna & $-11.5$ & $-23.0$ & $-8.7$ & $-19.5$ & $-3.1$ & $-3.6$\\
\modelSol & $-8.0$ & $-12.6$ & $-10.4$ & $-16.0$ & $-10.9$ & $-16.5$\\
\modelMimo & $+2.3$ & $-11.5$ & $-0.1$ & $-12.5$ & $-1.7$ & $-6.0$\\
\modelQwen & $-9.2$ & $-17.2$ & $-4.0$ & $-11.1$ & $-8.1$ & $-10.1$\\
\modelHaiku & $-2.3$ & $-12.6$ & $-1.9$ & $-6.9$ & $-2.3$ & $-17.4$\\
\modelQwenFlash & $-13.8$ & $-10.3$ & $-6.3$ & $-22.4$ & $-3.4$ & $-13.0$\\
\midrule
\rowcolor{hlRow}Average & $-7.1$ & $-14.6$ & $-5.2$ & $-14.7$ & $-4.9$ & $-11.1$\\
\bottomrule
\end{tabular}
\caption{\textbf{By round 30, every agent loses pass rate on every benchmark.} Change in pass rate or accuracy from the original skill after rounds 1 and 30. R1 and R30 denote these rounds; pp denotes percentage points.}
\label{tab:decline}
\end{table}

\begin{table}[t]
\centering\footnotesize
\setlength{\tabcolsep}{2pt}
\resizebox{\columnwidth}{!}{\fitcols{3}{MedCalc-Bench}{$-0.32$}
\begin{tabular}{@{}lWW>{\columncolor{hlCol}}WWW>{\columncolor{hlCol}}WWW>{\columncolor{hlCol}}W@{}}
\toprule
& \multicolumn{3}{c}{SkillsBench} & \multicolumn{3}{c}{LogicBench} & \multicolumn{3}{c}{MedCalc-Bench}\\
\cmidrule(lr){2-4}\cmidrule(lr){5-7}\cmidrule(l){8-10}
Model & Orig. & R30 & $\Delta$ & Orig. & R30 & $\Delta$ & Orig. & R30 & $\Delta$\\
\midrule
\modelLuna & 1.50 & 1.18 & $-0.32$ & 1.38 & 1.23 & $-0.15$ & 1.36 & 1.18 & $-0.18$\\
\modelSol & 1.68 & 1.34 & $-0.34$ & 1.43 & 1.22 & $-0.21$ & 1.68 & 1.39 & $-0.29$\\
\modelMimo & 1.29 & 1.18 & $-0.11$ & 1.21 & 1.07 & $-0.14$ & 1.22 & 1.10 & $-0.12$\\
\modelQwen & 1.41 & 1.28 & $-0.13$ & 1.32 & 1.18 & $-0.14$ & 1.31 & 1.23 & $-0.08$\\
\modelHaiku & 1.45 & 1.35 & $-0.10$ & 1.25 & 1.22 & $-0.03$ & 1.41 & 1.36 & $-0.05$\\
\modelQwenFlash & 1.33 & 1.20 & $-0.13$ & 1.21 & 1.16 & $-0.05$ & 1.22 & 1.17 & $-0.05$\\
\midrule
\rowcolor{hlRow}Average & 1.44 & 1.26 & $-0.18$ & 1.30 & 1.18 & $-0.12$ & 1.37 & 1.24 & $-0.13$\\
\bottomrule
\end{tabular}}
\caption{\textbf{On every benchmark, rewriting brings every model's skills closer to its self-written skills.} Distance of the original skill and of the round-30 skill to the model's ten self-written skills over TF-IDF word vectors, in units of the mean distance between two self-written skills. Orig. and R30 denote the two skill versions; $\Delta$ denotes the change in distance between them.}
\label{tab:distance}
\end{table}

\subsection{Loss of Items by Model, Benchmark and Kind}
\label{sec:suppl_loss}

\Cref{tab:suppl_loss} gives, for every model on every benchmark, the share of items lost after rounds 1 and 30, separately for the items outside the prior and for the other items.
Every model loses the items outside the prior more often on every benchmark, by 24 points on average over models and benchmarks already after the first rewrite and by 41 points after thirty rounds, when 77\% of the items outside the prior are lost against 35\% of the other items.
\Cref{tab:suppl_kindloss} breaks the loss on SkillsBench down by the kind of item, leaving out tables and front matter, of which a skill holds too few to compare.
Within list entries, paragraphs, code blocks and headings alike, every model loses the items outside the prior more often, by 36 points on average over the models and kinds, so the selective loss does not come from the kinds of content that a rewrite tends to shorten.

\begin{table*}[t]
\centering\footnotesize
\setlength{\tabcolsep}{5pt}
\fitcols{3}{MedCalc-Bench}{$+42.1$}
\begin{tabular}{@{}lWW>{\columncolor{hlCol}}WWW>{\columncolor{hlCol}}WWW>{\columncolor{hlCol}}W@{}}
\toprule
& \multicolumn{3}{c}{SkillsBench} & \multicolumn{3}{c}{LogicBench} & \multicolumn{3}{c}{MedCalc-Bench}\\
\cmidrule(lr){2-4}\cmidrule(lr){5-7}\cmidrule(l){8-10}
Model & 0/10 & $\geq$1/10 & $\Delta$ & 0/10 & $\geq$1/10 & $\Delta$ & 0/10 & $\geq$1/10 & $\Delta$\\
\midrule
\multicolumn{10}{@{}l}{\textit{After round 1}}\\
\modelLuna & 73.1 & 31.6 & $+41.5$ & 39.4 & 15.6 & $+23.8$ & 45.5 & 21.8 & $+23.7$\\
\modelSol & 75.4 & 33.3 & $+42.1$ & 40.9 & 16.6 & $+24.3$ & 57.5 & 18.9 & $+38.6$\\
\modelMimo & 23.1 & 11.9 & $+11.2$ & 25.7 & 4.3 & $+21.4$ & 26.0 & 8.7 & $+17.3$\\
\modelQwen & 37.5 & 20.1 & $+17.4$ & 38.3 & 8.7 & $+29.6$ & 30.6 & 13.2 & $+17.4$\\
\modelHaiku & 29.6 & 16.5 & $+13.1$ & 20.4 & 5.8 & $+14.6$ & 20.1 & 9.7 & $+10.4$\\
\modelQwenFlash & 58.2 & 24.8 & $+33.4$ & 35.5 & 14.9 & $+20.6$ & 36.3 & 12.1 & $+24.2$\\
Average & 49.5 & 23.0 & $+26.4$ & 33.4 & 11.0 & $+22.4$ & 36.0 & 14.1 & $+21.9$\\
\midrule
\multicolumn{10}{@{}l}{\textit{After round 30}}\\
\modelLuna & 90.0 & 42.6 & $+47.4$ & 82.4 & 40.2 & $+42.2$ & 80.8 & 38.7 & $+42.1$\\
\modelSol & 93.4 & 48.2 & $+45.2$ & 88.6 & 37.1 & $+51.5$ & 82.2 & 34.0 & $+48.2$\\
\modelMimo & 68.5 & 29.1 & $+39.4$ & 75.1 & 23.8 & $+51.3$ & 78.0 & 32.4 & $+45.6$\\
\modelQwen & 79.1 & 39.3 & $+39.8$ & 81.9 & 29.2 & $+52.7$ & 58.9 & 25.1 & $+33.8$\\
\modelHaiku & 62.9 & 34.3 & $+28.6$ & 43.6 & 14.4 & $+29.2$ & 48.2 & 24.1 & $+24.1$\\
\modelQwenFlash & 94.1 & 54.0 & $+40.1$ & 82.8 & 44.5 & $+38.3$ & 93.5 & 47.7 & $+45.8$\\
Average & 81.3 & 41.2 & $+40.1$ & 75.7 & 31.5 & $+44.2$ & 73.6 & 33.7 & $+39.9$\\
\bottomrule
\end{tabular}
\caption{\textbf{After one round and after thirty, every model loses the items outside the prior more often than the other items, on every benchmark.} Share of items lost, in percent, for the 0/10 items outside the prior, which none of the model's ten self-written skills states, and for the $\geq$1/10 items that some self-written skill states; $\Delta$ is the difference between them in pp.}
\label{tab:suppl_loss}
\end{table*}

\begin{table*}[t]
\centering\footnotesize
\setlength{\tabcolsep}{4.5pt}
\fitcols{3}{List entries}{$+47.7$}
\begin{tabular}{@{}lWW>{\columncolor{hlCol}}WWW>{\columncolor{hlCol}}WWW>{\columncolor{hlCol}}WWW>{\columncolor{hlCol}}W@{}}
\toprule
& \multicolumn{3}{c}{List entries} & \multicolumn{3}{c}{Paragraphs} & \multicolumn{3}{c}{Code blocks} & \multicolumn{3}{c}{Headings}\\
\cmidrule(lr){2-4}\cmidrule(lr){5-7}\cmidrule(lr){8-10}\cmidrule(lr){11-13}
Model & 0/10 & $\geq$1/10 & $\Delta$ & 0/10 & $\geq$1/10 & $\Delta$ & 0/10 & $\geq$1/10 & $\Delta$ & 0/10 & $\geq$1/10 & $\Delta$\\
\midrule
\modelLuna & 90.3 & 42.6 & $+47.7$ & 89.0 & 41.7 & $+47.3$ & 92.4 & 59.5 & $+32.9$ & 85.5 & 40.3 & $+45.2$\\
\modelSol & 94.5 & 47.2 & $+47.3$ & 93.3 & 49.6 & $+43.7$ & 90.4 & 57.4 & $+33.0$ & 92.4 & 47.2 & $+45.2$\\
\modelMimo & 68.5 & 33.5 & $+35.0$ & 68.5 & 26.9 & $+41.6$ & 73.2 & 30.2 & $+43.0$ & 50.6 & 23.3 & $+27.3$\\
\modelQwen & 80.8 & 45.2 & $+35.6$ & 79.5 & 41.3 & $+38.2$ & 81.5 & 52.8 & $+28.7$ & 55.0 & 30.0 & $+25.0$\\
\modelHaiku & 59.3 & 34.2 & $+25.1$ & 65.1 & 36.5 & $+28.6$ & 73.5 & 51.7 & $+21.8$ & 49.0 & 29.2 & $+19.8$\\
\modelQwenFlash & 93.9 & 54.1 & $+39.8$ & 92.3 & 55.0 & $+37.3$ & 96.6 & 66.7 & $+29.9$ & 92.1 & 50.9 & $+41.2$\\
\midrule
Average & 81.2 & 42.8 & $+38.4$ & 81.3 & 41.8 & $+39.4$ & 84.6 & 53.1 & $+31.6$ & 70.8 & 36.8 & $+33.9$\\
\bottomrule
\end{tabular}
\caption{\textbf{Within list entries, paragraphs, code blocks and headings alike, every model loses the items outside the prior more often.} Share of the items of each kind lost after thirty rounds on SkillsBench, in percent, for the 0/10 items that none of the model's ten self-written skills states and for the other items in the $\geq$1/10 group; $\Delta$ is the difference between them in pp.}
\label{tab:suppl_kindloss}
\end{table*}

\subsection{Reliability of the Judge}
\label{sec:suppl_judge}

The analysis of the items rests on the labels of one LLM judge, so we check whether these labels, and the gap between the items outside the prior and the other items, depend on the judge or on the run.
We draw 20 SkillsBench tasks with a fixed seed, which hold 1,800 items, and judge them again in three ways with the prompts and settings of the main text.
\modelSol, as a second judge, labels the items against the ten self-written skills of \modelLuna and against its rewritten skills of rounds 1, 5, 10 and 30, \modelHaiku, from another family of models, labels them as a third judge, and \modelLuna labels them a second time.
Each judgment forms the two classes of items, outside the prior or not, from its own labels on the self-written skills.

\begin{table}[t]
\centering\footnotesize
\setlength{\tabcolsep}{3pt}
\resizebox{\columnwidth}{!}{\begin{tabular}{@{}lccc@{}}
\toprule
& Second judge & Main judge, rerun & Third judge\\
& \modelSol & \modelLuna & \modelHaiku\\
\midrule
Agreement on self-written skills, \% & 89.7 & 92.7 & 87.2\\
\quad Cohen's $\kappa$ & 0.78 & 0.84 & 0.72\\
Agreement on 0/10 class, \% & 88.9 & 94.8 & 86.0\\
Agreement on rewritten skills, \% & 87.2 & 89.8 & 82.4\\
\quad Cohen's $\kappa$ & 0.74 & 0.79 & 0.64\\
\bottomrule
\end{tabular}}
\caption{\textbf{Two further judges and a second run agree with the judgments of the main text.} Share of the labels on which each agrees with the main judgments, in percent, and Cohen's $\kappa$, on 20 SkillsBench tasks with 1,800 items. An item is in the 0/10 class when none of the ten self-written skills states it.}
\label{tab:suppl_judgerel}
\end{table}

\begin{table}[t]
\centering\footnotesize
\setlength{\tabcolsep}{3pt}
\resizebox{\columnwidth}{!}{\begin{tabular}{@{}lcccc@{}}
\toprule
& Main judge & Second judge & Main judge, rerun & Third judge\\
& \modelLuna & \modelSol & \modelLuna & \modelHaiku\\
\midrule
0/10 & 89.2 & 93.3 & 89.9 & 93.4\\
$\geq$1/10 & 39.8 & 40.9 & 38.4 & 52.6\\
\rowcolor{hlRow}$\Delta$, pp & $+49.5$ & $+52.5$ & $+51.4$ & $+40.7$\\
\bottomrule
\end{tabular}}
\caption{\textbf{Every judgment finds a large gap.} Share of items lost after thirty rounds, in percent, for the 0/10 items outside the prior and for the other items in the $\geq$1/10 group, on the same 20 SkillsBench tasks under each judgment; $\Delta$ is the difference in pp.}
\label{tab:suppl_judgegap}
\end{table}

\paragraph{The labels agree.}
The second judge gives the same label as the main judgments on 89.7\% of the labels on the self-written skills and on 87.2\% of the kept or lost labels on the rewritten skills, with Cohen's $\kappa$ of 0.78 and 0.74, and it places 88.9\% of the items in the same class (\cref{tab:suppl_judgerel}).
The second run of \modelLuna agrees even more closely, on 92.7\% and 89.8\% of the labels and on 94.8\% of the classes.
The third judge agrees on 87.2\% and 82.4\% of the labels, with $\kappa$ of 0.72 and 0.64, and places 86.0\% of the items in the same class.

\paragraph{The gap does not depend on the judge.}
After thirty rounds, the items outside the prior are lost 49.5 points more often than the other items under the main judgments on these tasks, 52.5 points more often under the second judge, 51.4 points more often under the second run and 40.7 points more often under the third judge, which labels more of the other items lost as well (\cref{tab:suppl_judgegap}).

\paragraph{The judge finds no loss where there is none.}
Two controls have known answers.
When the round-30 skill is replaced by the original word for word, every item should be kept, and neither \modelLuna nor \modelSol labels any item lost.
When the round-1 skill is replaced by the original skill of another task, every item should be lost, and the two label only 1.3\% and 1.5\% of the items kept.
The selective loss in the main text is thus not an artifact of the judge or of a single run.

\subsection{Forty Self-Written Skills}
\label{sec:suppl_forty}

Whether an item lies outside the prior is decided by ten self-written skills, and an item that the model writes only now and then could be missing from all ten by chance.
For \modelLuna on SkillsBench, the model therefore writes thirty further self-written skills for every skill with the prompt and settings of the first ten, and the judge labels every item against them, which gives three further samples of ten.

\begin{table}[t]
\centering\footnotesize
\begin{tabular}{@{}lcc@{}}
\toprule
Samples of ten & Agreement on 0/10 class, \% & Cohen's $\kappa$\\
\midrule
Original vs.\ further 1 & 88.1 & 0.76\\
Original vs.\ further 2 & 88.2 & 0.76\\
Original vs.\ further 3 & 87.7 & 0.75\\
Further 1 vs.\ 2 & 92.0 & 0.84\\
Further 1 vs.\ 3 & 91.7 & 0.83\\
Further 2 vs.\ 3 & 91.4 & 0.83\\
\bottomrule
\end{tabular}
\caption{\textbf{Every sample of ten self-written skills finds nearly the same items outside the prior.} Share of the 6,605 SkillsBench items that two samples of ten self-written skills of \modelLuna place in the same class, outside the prior or not, and Cohen's $\kappa$. The original sample is the one of the main text.}
\label{tab:suppl_forty_agree}
\end{table}

\begin{table}[t]
\centering\footnotesize
\begin{tabular}{@{}lcc@{}}
\toprule
Class under the original ten & Items & Stated by the further thirty, \%\\
\midrule
0/10 & 2,757 & 1.1\\
1--9/10 & 2,584 & 39.6\\
10/10 & 1,264 & 95.1\\
\bottomrule
\end{tabular}
\caption{\textbf{Items outside the prior are almost never written by further self-written skills.} SkillsBench items of \modelLuna by how many of the original ten self-written skills state them, and the mean share of the thirty further self-written skills that state an item.}
\label{tab:suppl_forty_rate}
\end{table}

\paragraph{Every sample of ten finds nearly the same items.}
Each sample of ten forms its own class of items outside the prior.
The original sample places 88\% of the items in the same class as each further sample on average, with Cohen's $\kappa$ of 0.76, and the further samples agree with one another on 92\% of the items, with $\kappa$ of 0.83 (\cref{tab:suppl_forty_agree}).

\paragraph{Items outside the prior are almost never written.}
The thirty further self-written skills play no part in forming the class of the original ten, so they show how often the model writes each item on its own.
Even in these thirty further samples, an item in the original 0/10 group appears in only 1.1\% of the documents on average, while an item in the original 10/10 group appears in 95.1\% (\cref{tab:suppl_forty_rate}).
The sharp separation thus persists in new samples that played no part in the classification.

\subsection{How Items Are Lost and Whether They Return}
\label{sec:suppl_losstype}

We use the SkillsBench item judgments to classify the losses at round thirty and check whether items lost in the first rewrite appear again in the round-thirty skills.
Among the items lost after thirty rounds on SkillsBench, those outside the prior are on average dropped outright more often than the other lost items, 53\% against 40\%, and weakened less often, 29\% against 42\%, while a similar share of both is changed to different content (\cref{tab:suppl_losstype}).
A lost item also rarely returns when it lies outside the prior.
Of the items that a model loses in its first rewrite, only 6\% of those outside the prior are kept again after thirty rounds on average, against 30\% of the other items (\cref{tab:suppl_regain}).
A rewrite thus tends to remove the content outside the prior entirely, and the model does not write it back, whereas lost content within the prior can return in a later round.

\begin{table}[t]
\centering\footnotesize
\setlength{\tabcolsep}{4pt}
\fitcols{2}{Dropped outright}{$\geq$1/10}
\begin{tabular}{@{}l>{\columncolor{hlAuth}}WW>{\columncolor{hlAuth}}WW@{}}
\toprule
& \multicolumn{2}{c}{Dropped outright} & \multicolumn{2}{c}{Weakened}\\
\cmidrule(lr){2-3}\cmidrule(l){4-5}
Model & 0/10 & $\geq$1/10 & 0/10 & $\geq$1/10\\
\midrule
\modelLuna & 55 & 37 & 28 & 48\\
\modelSol & 60 & 39 & 24 & 43\\
\modelMimo & 47 & 40 & 32 & 42\\
\modelQwen & 52 & 43 & 27 & 39\\
\modelHaiku & 40 & 41 & 38 & 42\\
\modelQwenFlash & 63 & 43 & 24 & 39\\
\midrule
\rowcolor{hlRow}Average & 53 & 40 & 29 & 42\\
\bottomrule
\end{tabular}
\caption{\textbf{On average, lost items outside the prior are dropped outright more often, and weakened less often, than the other lost items.} Of the items lost after thirty rounds on SkillsBench, the share dropped outright and the share weakened, in percent of the lost items, for the 0/10 items outside the prior and for the other items in the $\geq$1/10 group. The remaining lost items are changed to different content.}
\label{tab:suppl_losstype}
\end{table}

\begin{table}[t]
\centering\footnotesize
\begin{tabular}{@{}l>{\columncolor{hlAuth}}cc@{}}
\toprule
Model & 0/10 & $\geq$1/10\\
\midrule
\modelLuna & 3 & 26\\
\modelSol & 2 & 21\\
\modelMimo & 10 & 42\\
\modelQwen & 6 & 29\\
\modelHaiku & 12 & 38\\
\modelQwenFlash & 2 & 26\\
\midrule
\rowcolor{hlRow}Average & 6 & 30\\
\bottomrule
\end{tabular}
\caption{\textbf{Lost items outside the prior rarely return.} Of the SkillsBench items lost in the first rewrite, the share kept again after thirty rounds, in percent, for the 0/10 items outside the prior and for the other items in the $\geq$1/10 group.}
\label{tab:suppl_regain}
\end{table}

\subsection{Overlap of the Items Outside the Prior across Models}
\label{sec:suppl_overlap}

Whether the content outside the prior is particular to one model or shared by many bears on what the rewriting gives up.
For every pair of models, we take the SkillsBench items outside the prior of one model and count how many of them lie outside the prior of the other model as well.
If the two sets were unrelated, this share would equal the share of all items that lie outside the prior of the other model.
The observed share is far higher, 82\% on average against 48\% expected (\cref{tab:suppl_overlap}), no single pair of models falls below 70\%, and 1,866 of the 6,605 items lie outside the prior of every model.
The knowledge outside the prior is thus largely knowledge that the models in general do not write on their own rather than a gap of one model, and every model that rewrites a skill tends to give up the same part of it.

\begin{table}[t]
\centering\footnotesize
\setlength{\tabcolsep}{2.5pt}
\fitcols{2}{Also 0/10 for another model, \%}{Expected}
\begin{tabular}{@{}lc>{\columncolor{hlAuth}}WW@{}}
\toprule
& 0/10 & \multicolumn{2}{c}{Also 0/10 for another model, \%}\\
\cmidrule(l){3-4}
Model & items & Observed & Expected\\
\midrule
\modelLuna & 2,757 & 87 & 49\\
\modelSol & 3,019 & 84 & 48\\
\modelMimo & 3,419 & 78 & 47\\
\modelQwen & 3,410 & 78 & 47\\
\modelHaiku & 2,924 & 84 & 48\\
\modelQwenFlash & 3,333 & 80 & 47\\
\midrule
\rowcolor{hlRow}Average & -- & 82 & 48\\
\bottomrule
\end{tabular}
\caption{\textbf{The items outside the prior are largely the same for every model.} For the SkillsBench items outside the prior of each model, the share that also lies outside the prior of another model, averaged over the other models, against the share expected if the two sets were unrelated, which is the share of all items that lie outside the other model's prior.}
\label{tab:suppl_overlap}
\end{table}

\subsection{The Distance Measure}
\label{sec:suppl_distance}

The distance to the prior in the main text uses one measure, TF-IDF word vectors on a scale set by the self-written skills.
We check two things, that this scale means what it is meant to mean, and that other measures show the same movement of the rewritten skills.

\begin{table}[t]
\centering\footnotesize
\setlength{\tabcolsep}{2.6pt}
\resizebox{\columnwidth}{!}{\fitcols{2}{Pairs closer than the original}{Round 30}
\begin{tabular}{@{}lccccWW@{}}
\toprule
& Held-out & \multicolumn{3}{c}{Distance} & \multicolumn{2}{c}{Pairs closer than the original}\\
\cmidrule(lr){3-5}\cmidrule(l){6-7}
Measure & self-written skill & Original & Round 5 & Round 30 & Round 5 & Round 30\\
\midrule
\rowcolor{hlRow}TF-IDF in main text & 1.00 & 1.37 & 1.19 & 1.22 & 18/18 & 18/18\\
Embedding & 1.01 & 2.24 & 1.78 & 1.93 & 17/18 & 15/18\\
Word frequency & 1.00 & 2.42 & 1.62 & 1.77 & 18/18 & 16/18\\
\bottomrule
\end{tabular}}
\caption{\textbf{Every measure places a held-out self-written skill at 1 and moves the rewritten skills toward the self-written skills in the first rounds.} Distance to the model's ten self-written skills, in units of the mean distance between two self-written skills, averaged over the 18 pairs of model and benchmark, and the number of pairs in which the rewritten skill lies closer than the original. The held-out self-written skill is measured against the other nine on the same scale.}
\label{tab:suppl_measures}
\end{table}

\input{fig/distance_emb}
\input{fig/distance_tf}

\paragraph{The scale is calibrated.}
On the scale of the main text, a skill at 1 lies as close to the self-written skills as they lie to one another.
To test this, we hold out each self-written skill in turn and measure it against the other nine on the same scale.
It lies at 1.00 under TF-IDF for every model and benchmark, and at 1.01 on average under the two other measures below (\cref{tab:suppl_measures}), so a further sample of the prior indeed lies at 1.
The original skills lie above 1 under every measure, so each measure separates the expert's text from what the model writes.

\paragraph{Other measures show the same movement.}
We repeat the measurement with two other representations of a document, Qwen3-Embedding vectors~\cite{zhang_2025_qwen3embedding}, which represent the meaning of a document rather than its exact wording, and raw word frequencies without the TF-IDF weights.
Under both, rewriting moves the skills toward the self-written skills over the first rounds, after which the distance levels off (\cref{tab:suppl_measures}), and \cref{fig:suppl_distance_emb,fig:suppl_distance_tf} show this for every model and benchmark in the layout of \cref{fig:distance}.
On average over the 18 pairs of model and benchmark, the distance falls from 2.24 to 1.78 by round five under embeddings and from 2.42 to 1.62 under word frequencies, and it changes little afterwards.
By round five, the rewritten skill lies closer than the original in every pair under word frequencies and in all pairs but one under embeddings.
We keep TF-IDF in the main text because the contribution of an expert often lies in specific commands, paths and parameter names, which TF-IDF weighs heavily, whereas embeddings weigh the topic and style that every skill on a topic shares, and raw frequencies are carried by common words.

\FloatBarrier
\section{Additional Experiments}
\label{sec:suppl_experiments}

\subsection{\texorpdfstring{$\tau^2$}{tau2}-bench}
\label{sec:suppl_tau2}

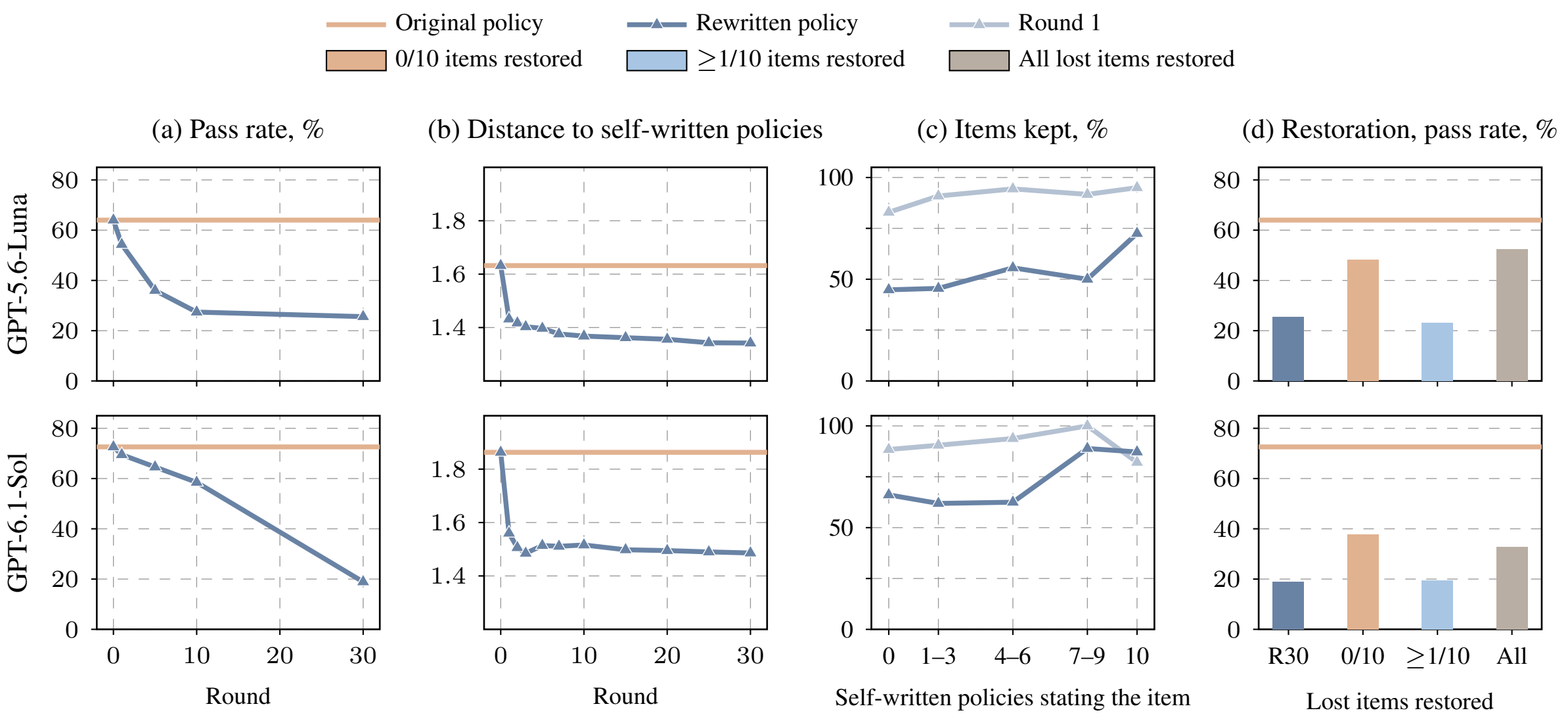
\begin{figure*}[t]
\centering
\begin{tikzpicture}
\begin{groupplot}[group style={group size=4 by 2, horizontal sep=1.05cm, vertical sep=0.35cm}, scale only axis, width=2.85cm, height=2.15cm, axis lines=box, axis line style={black, line width=0.5pt}, tick style={black, line width=0.4pt}, major tick length=1.6pt, tick align=outside, xtick pos=bottom, ytick pos=left, xmajorgrids, ymajorgrids, major grid style={black!40, dashed, line width=0.25pt}, tick label style={font=\scriptsize}, label style={font=\scriptsize}, title style={font=\footnotesize, yshift=-3pt, text height=1.6ex, text depth=0.4ex}]
\nextgroupplot[xmin=-2, xmax=32, xtick={0,10,20,30}, ymin=0, ymax=85, ytick={0,20,40,60,80}, title={(a) Pass rate, \%}, ylabel={{\footnotesize \modelLuna}}, xticklabels={}, legend to name=legtau, legend columns=3, legend cell align=left, legend style={font=\scriptsize, draw=none, /tikz/every even column/.append style={column sep=10pt}}]
\addplot[cOrig, line width=1.4pt, domain=-2:32, samples=2, forget plot] {64.0};
\addplot[cRew, line width=1.4pt, mark=triangle*, mark size=2.2pt, mark options={solid, fill=cRew, draw=white, line width=0.25pt}, forget plot] coordinates {(0,64.0) (1,54.3) (5,36.0) (10,27.4) (30,25.6)};
\addlegendimage{cOrig, line width=1.4pt}\addlegendentry{Original policy}
\addlegendimage{cRew, line width=1.4pt, mark=triangle*, mark size=2.2pt, mark options={solid, fill=cRew, draw=white, line width=0.25pt}}\addlegendentry{Rewritten policy}
\addlegendimage{cRew!50, line width=1.4pt, mark=triangle*, mark size=2.2pt, mark options={solid, fill=cRew!50, draw=white, line width=0.25pt}}\addlegendentry{Round 1}
\addlegendimage{area legend, fill=cNone, draw=none}\addlegendentry{0/10 items restored}
\addlegendimage{area legend, fill=cSome, draw=none}\addlegendentry{$\geq$1/10 items restored}
\addlegendimage{area legend, fill=cAll, draw=none}\addlegendentry{All lost items restored}
\nextgroupplot[xmin=-2, xmax=32, xtick={0,10,20,30}, ymin=1.2, ymax=2.0, ytick={1.4,1.6,1.8}, yticklabel style={/pgf/number format/fixed, /pgf/number format/precision=1}, title={(b) Distance to self-written policies}, xticklabels={}]
\addplot[cOrig, line width=1.4pt, domain=-2:32, samples=2, forget plot] {1.632};
\addplot[cRew, line width=1.4pt, mark=triangle*, mark size=2.2pt, mark options={solid, fill=cRew, draw=white, line width=0.25pt}, forget plot] coordinates {(0,1.632) (1,1.432) (2,1.417) (3,1.403) (5,1.397) (7,1.376) (10,1.368) (15,1.362) (20,1.356) (25,1.343) (30,1.342)};
\nextgroupplot[xmin=-0.7, xmax=10.7, xtick={0,2,5,8,10}, xticklabels={0,1--3,4--6,7--9,10}, ymin=0, ymax=105, ytick={0,25,...,100}, yticklabels={0,,50,,100}, title={(c) Items kept, \%}, xticklabels={}]
\addplot[cRew!50, line width=1.4pt, mark=triangle*, mark size=2.2pt, mark options={solid, fill=cRew!50, draw=white, line width=0.25pt}, forget plot] coordinates {(0,82.9) (2,90.9) (5,94.4) (8,91.7) (10,95.0)};
\addplot[cRew, line width=1.4pt, mark=triangle*, mark size=2.2pt, mark options={solid, fill=cRew, draw=white, line width=0.25pt}, forget plot] coordinates {(0,44.8) (2,45.5) (5,55.6) (8,50.0) (10,72.5)};
\nextgroupplot[xmin=-0.4, xmax=3.4, xtick={0,1,2,3}, xticklabels={R30,0/10,$\geq$1/10,All}, xmajorgrids=false, ymin=0, ymax=85, ytick={0,20,40,60,80}, title={(d) Restoration, pass rate, \%}, xticklabels={}]
\addplot[cOrig, line width=1.4pt, domain=-0.4:3.4, samples=2, forget plot] {64.0};
\addplot[ybar, bar width=9pt, fill=cRew, draw=none, forget plot] coordinates {(0,25.6)};
\addplot[ybar, bar width=9pt, fill=cNone, draw=none, forget plot] coordinates {(1,48.2)};
\addplot[ybar, bar width=9pt, fill=cSome, draw=none, forget plot] coordinates {(2,23.2)};
\addplot[ybar, bar width=9pt, fill=cAll, draw=none, forget plot] coordinates {(3,52.4)};
\nextgroupplot[xmin=-2, xmax=32, xtick={0,10,20,30}, ymin=0, ymax=85, ytick={0,20,40,60,80}, ylabel={{\footnotesize \modelSol}}, xlabel={Round}]
\addplot[cOrig, line width=1.4pt, domain=-2:32, samples=2, forget plot] {72.6};
\addplot[cRew, line width=1.4pt, mark=triangle*, mark size=2.2pt, mark options={solid, fill=cRew, draw=white, line width=0.25pt}, forget plot] coordinates {(0,72.6) (1,69.5) (5,64.6) (10,58.5) (30,18.9)};
\nextgroupplot[xmin=-2, xmax=32, xtick={0,10,20,30}, ymin=1.2, ymax=2.0, ytick={1.4,1.6,1.8}, yticklabel style={/pgf/number format/fixed, /pgf/number format/precision=1}, xlabel={Round}]
\addplot[cOrig, line width=1.4pt, domain=-2:32, samples=2, forget plot] {1.863};
\addplot[cRew, line width=1.4pt, mark=triangle*, mark size=2.2pt, mark options={solid, fill=cRew, draw=white, line width=0.25pt}, forget plot] coordinates {(0,1.863) (1,1.56) (2,1.506) (3,1.485) (5,1.514) (7,1.511) (10,1.516) (15,1.498) (20,1.495) (25,1.49) (30,1.486)};
\nextgroupplot[xmin=-0.7, xmax=10.7, xtick={0,2,5,8,10}, xticklabels={0,1--3,4--6,7--9,10}, ymin=0, ymax=105, ytick={0,25,...,100}, yticklabels={0,,50,,100}, xlabel={Self-written policies stating the item}]
\addplot[cRew!50, line width=1.4pt, mark=triangle*, mark size=2.2pt, mark options={solid, fill=cRew!50, draw=white, line width=0.25pt}, forget plot] coordinates {(0,88.4) (2,90.5) (5,93.8) (8,100.0) (10,82.1)};
\addplot[cRew, line width=1.4pt, mark=triangle*, mark size=2.2pt, mark options={solid, fill=cRew, draw=white, line width=0.25pt}, forget plot] coordinates {(0,66.1) (2,61.9) (5,62.5) (8,88.9) (10,87.2)};
\nextgroupplot[xmin=-0.4, xmax=3.4, xtick={0,1,2,3}, xticklabels={R30,0/10,$\geq$1/10,All}, xmajorgrids=false, ymin=0, ymax=85, ytick={0,20,40,60,80}, xlabel={Lost items restored}]
\addplot[cOrig, line width=1.4pt, domain=-0.4:3.4, samples=2, forget plot] {72.6};
\addplot[ybar, bar width=9pt, fill=cRew, draw=none, forget plot] coordinates {(0,18.9)};
\addplot[ybar, bar width=9pt, fill=cNone, draw=none, forget plot] coordinates {(1,37.8)};
\addplot[ybar, bar width=9pt, fill=cSome, draw=none, forget plot] coordinates {(2,19.5)};
\addplot[ybar, bar width=9pt, fill=cAll, draw=none, forget plot] coordinates {(3,32.9)};
\end{groupplot}
\node[anchor=south, yshift=1pt] at (current bounding box.north) {\pgfplotslegendfromname{legtau}};
\end{tikzpicture}
\caption{\textbf{On $\tau^2$-bench, the analyses of the main text give the same picture for both models.} (a) Pass rate over the 164 tasks of the airline and retail domains with the original policy and after each round of rewriting. (b) TF-IDF distance of the policy to the model's ten self-written policies, in units of the mean distance between two self-written policies. (c) Share of the items of the original policies kept after rounds 1 and 30, by the number of self-written policies that state the item. (d) Pass rate of the round-30 policy and of the round-30 policy with the lost 0/10 items outside the prior, the other lost items in the $\geq$1/10 group or all lost items restored.}
\label{fig:suppl_tau2}
\end{figure*}

Beyond the three benchmarks of the main text, we verify the findings with \modelLuna and \modelSol on $\tau^2$-bench~\cite{barres_2025_tau2}, where what guides the agent is not a skill but a written policy.
The benchmark gives a conversational agent the rules of an airline or a retail service, and we apply the procedure of the main text to these two policies.
Each model rewrites each policy for thirty rounds, writes ten policies of its own from the domain's topic, and runs as the agent on the 164 tasks of the two domains.
We also build the three restored versions of each policy following the procedure in the main text and evaluate their pass rates on the same 164 tasks.
\Cref{fig:suppl_tau2} repeats the four analyses of the main text for both models.

\paragraph{Rewriting lowers the pass rate.}
Rewriting also causes large performance losses with the service policies of $\tau^2$-bench.
After thirty rounds, the pass rate falls from 64.0\% to 25.6\% for \modelLuna and from 72.6\% to 18.9\% for \modelSol (\cref{fig:suppl_tau2}a).

\paragraph{The policies move toward the self-written policies.}
The distance to the self-written policies falls from 1.63 to 1.34 for \modelLuna and from 1.86 to 1.49 for \modelSol, most of it over the first rounds (\cref{fig:suppl_tau2}b).

\paragraph{The items outside the prior are lost more often.}
After thirty rounds, \modelLuna keeps 45\% of the items outside the prior against 73\% of the items that all self-written policies state, and \modelSol keeps 66\% against 87\% (\cref{fig:suppl_tau2}c).

\paragraph{Putting these items back recovers more of the pass rate.}
Putting the lost items outside the prior back into the round-30 policy lifts the pass rate to 48.2\% for \modelLuna and 37.8\% for \modelSol, whereas putting back the other lost items leaves it at 23.2\% and 19.5\% (\cref{fig:suppl_tau2}d).

\subsection{Rewriting Without Web Search}
\label{sec:suppl_web}

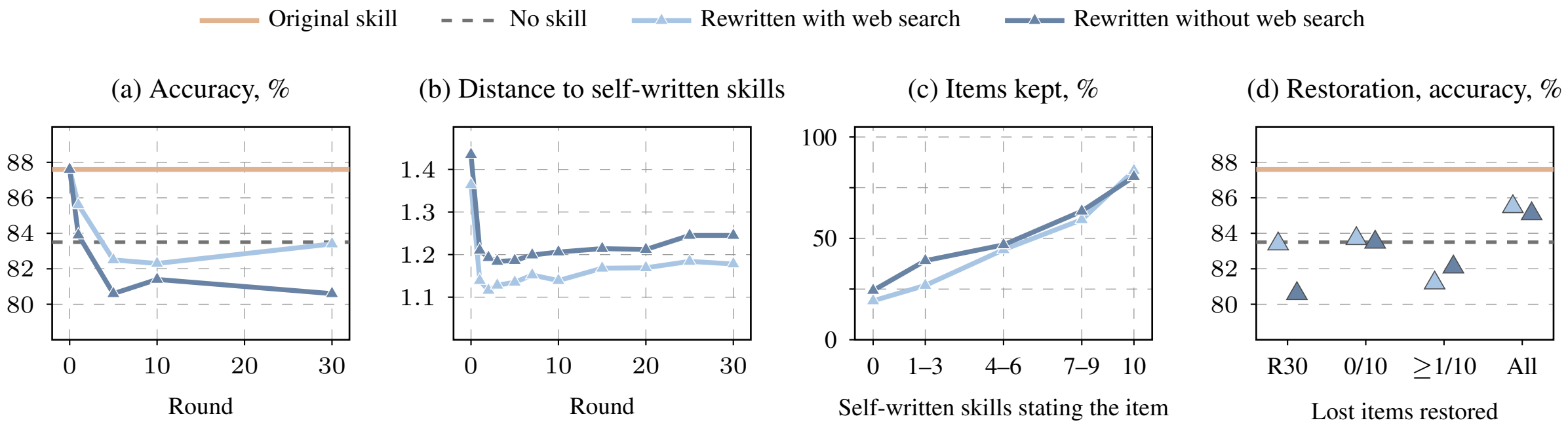
\begin{figure*}[t]
\centering
\begin{tikzpicture}
\begin{groupplot}[group style={group size=4 by 1, horizontal sep=1.05cm, vertical sep=0.35cm}, scale only axis, width=3.0cm, height=2.15cm, axis lines=box, axis line style={black, line width=0.5pt}, tick style={black, line width=0.4pt}, major tick length=1.6pt, tick align=outside, xtick pos=bottom, ytick pos=left, xmajorgrids, ymajorgrids, major grid style={black!40, dashed, line width=0.25pt}, tick label style={font=\scriptsize}, label style={font=\scriptsize}, title style={font=\footnotesize, yshift=-3pt, text height=1.6ex, text depth=0.4ex}]
\nextgroupplot[xmin=-2, xmax=32, xtick={0,10,20,30}, ymin=78, ymax=90, ytick={80,82,84,86,88}, title={(a) Accuracy, \%}, xlabel={Round}, legend to name=legweb, legend columns=-1, legend cell align=left, legend style={font=\scriptsize, draw=none, /tikz/every even column/.append style={column sep=10pt}}]
\addplot[cOrig, line width=1.4pt, domain=-2:32, samples=2, forget plot] {87.6};
\addplot[black!55, dashed, line width=1.0pt, domain=-2:32, samples=2, forget plot] {83.5};
\addplot[cSome, line width=1.4pt, mark=triangle*, mark size=2.2pt, mark options={solid, fill=cSome, draw=white, line width=0.25pt}, forget plot] coordinates {(0,87.6) (1,85.6) (5,82.5) (10,82.3) (30,83.4)};
\addplot[cRew, line width=1.4pt, mark=triangle*, mark size=2.2pt, mark options={solid, fill=cRew, draw=white, line width=0.25pt}, forget plot] coordinates {(0,87.6) (1,83.9) (5,80.6) (10,81.4) (30,80.6)};
\addlegendimage{cOrig, line width=1.4pt}\addlegendentry{Original skill}
\addlegendimage{black!55, dashed, line width=1.0pt}\addlegendentry{No skill}
\addlegendimage{cSome, line width=1.4pt, mark=triangle*, mark size=2.2pt, mark options={solid, fill=cSome, draw=white, line width=0.25pt}}\addlegendentry{Rewritten with web search}
\addlegendimage{cRew, line width=1.4pt, mark=triangle*, mark size=2.2pt, mark options={solid, fill=cRew, draw=white, line width=0.25pt}}\addlegendentry{Rewritten without web search}
\nextgroupplot[xmin=-2, xmax=32, xtick={0,10,20,30}, ymin=1.0, ymax=1.5, ytick={1.1,1.2,1.3,1.4}, yticklabel style={/pgf/number format/fixed, /pgf/number format/precision=1}, title={(b) Distance to self-written skills}, xlabel={Round}]
\addplot[cSome, line width=1.4pt, mark=triangle*, mark size=2.2pt, mark options={solid, fill=cSome, draw=white, line width=0.25pt}, forget plot] coordinates {(0,1.364) (1,1.139) (2,1.116) (3,1.128) (5,1.135) (7,1.152) (10,1.139) (15,1.168) (20,1.169) (25,1.184) (30,1.178)};
\addplot[cRew, line width=1.4pt, mark=triangle*, mark size=2.2pt, mark options={solid, fill=cRew, draw=white, line width=0.25pt}, forget plot] coordinates {(0,1.435) (1,1.21) (2,1.193) (3,1.184) (5,1.186) (7,1.199) (10,1.206) (15,1.214) (20,1.212) (25,1.245) (30,1.245)};
\nextgroupplot[xmin=-0.7, xmax=10.7, xtick={0,2,5,8,10}, xticklabels={0,1--3,4--6,7--9,10}, ymin=0, ymax=105, ytick={0,25,...,100}, yticklabels={0,,50,,100}, title={(c) Items kept, \%}, xlabel={Self-written skills stating the item}]
\addplot[cSome, line width=1.4pt, mark=triangle*, mark size=2.2pt, mark options={solid, fill=cSome, draw=white, line width=0.25pt}, forget plot] coordinates {(0,19.2) (2,26.8) (5,44.4) (8,59.2) (10,83.4)};
\addplot[cRew, line width=1.4pt, mark=triangle*, mark size=2.2pt, mark options={solid, fill=cRew, draw=white, line width=0.25pt}, forget plot] coordinates {(0,24.3) (2,39.0) (5,46.8) (8,63.4) (10,80.4)};
\nextgroupplot[xmin=-0.4, xmax=3.4, xtick={0,1,2,3}, xticklabels={R30,0/10,$\geq$1/10,All}, xmajorgrids=false, ymin=78, ymax=90, ytick={80,82,84,86,88}, title={(d) Restoration, accuracy, \%}, xlabel={Lost items restored}]
\addplot[cOrig, line width=1.4pt, domain=-0.4:3.4, samples=2, forget plot] {87.6};
\addplot[black!55, dashed, line width=1.0pt, domain=-0.4:3.4, samples=2, forget plot] {83.5};
\addplot[only marks, mark=triangle*, mark size=3.4pt, mark options={fill=cSome, draw=black!70, line width=0.3pt}, forget plot] coordinates {(-0.12,83.4) (0.88,83.7) (1.88,81.2) (2.88,85.5)};
\addplot[only marks, mark=triangle*, mark size=3.4pt, mark options={fill=cRew, draw=black!70, line width=0.3pt}, forget plot] coordinates {(0.12,80.6) (1.12,83.5) (2.12,82.1) (3.12,85.1)};
\end{groupplot}
\node[anchor=south, yshift=1pt] at (current bounding box.north) {\pgfplotslegendfromname{legweb}};
\end{tikzpicture}
\caption{\textbf{Without web search, rewriting lowers the accuracy of \modelLuna on MedCalc-Bench more, and the other analyses keep their shape.} (a) Accuracy on the 1,100 questions after each round. (b) TF-IDF distance to the ten self-written skills of the same version, written with or without web search. (c) Share of the items kept after thirty rounds, by the number of self-written skills that state the item. (d) Accuracy of the round-30 skill and of the round-30 skill with the lost 0/10 items outside the prior, the other lost items in the $\geq$1/10 group or all lost items restored. Both versions share the runs without a skill and with the original skill.}
\label{fig:suppl_web}
\end{figure*}

Codex can search the web while a model rewrites a skill or writes one from the topic, so content found online may enter both the rewritten and the self-written skills.
To see whether this changes the results, we repeat the MedCalc-Bench study for \modelLuna with web search turned off while it rewrites the skills and writes its self-written skills, and keep everything else, the 55 skills, thirty rounds, ten self-written skills per skill, the prompts, the length and the reasoning effort.
Both versions are evaluated on the 1,100 questions by \modelLuna in Codex at minimal reasoning with web search off, share the runs without a skill and with the original skill, and are compared in the four analyses of the main text (\cref{fig:suppl_web}).

\paragraph{Both versions show the findings of the main text.}
After thirty rounds, the skill rewritten without web search lies 7.0 points below the original skill and the skill rewritten with web search 4.3 points (\cref{fig:suppl_web}a).
Both move toward their self-written skills, mostly in the first round, to 1.25 and 1.18 units after thirty rounds (\cref{fig:suppl_web}b).
Both keep far fewer of the items outside the prior than of the items that all self-written skills state, 24\% against 80\% without web search and 19\% against 83\% with it (\cref{fig:suppl_web}c).
In both, putting back the lost items outside the prior recovers more than putting back the other lost items, +2.8 against +1.5 points without web search and +0.4 against $-$2.2 points with it (\cref{fig:suppl_web}d).

\paragraph{Without web search, the loss is larger.}
Without web search, what the model writes on its own lies further from what the author wrote, as the original skills lie 1.44 rather than 1.36 units from the self-written skills under TF-IDF and 2.19 rather than 1.59 under embeddings.
The rewritten skill then lies lower at every round, by 1.8 points on average, and putting back its lost items outside the prior recovers more.
Web search thus returns part of the content that the rewriting drops but does not stop the selective loss, and without it the rewriting relies more on the model's own idea of a skill and costs more.
The main text keeps the version with web search, the setting with the smaller loss.

\subsection{Repeated Runs of the Agent}
\label{sec:suppl_passk}

Every condition of the main text runs once per task, so a pass rate also carries the chance of a single run.
To see how the loss behaves when the agent has more attempts, \modelLuna runs every SkillsBench task at least sixteen times with the original skill and sixteen times with the round-30 skill.
From these runs we estimate pass@$k$, the chance that at least one of $k$ runs passes a task, with the unbiased estimator of \citet{chen_2021_codex} for $k$ from 1 to 16.
At large $k$, pass@$k$ shows which tasks an agent solves at least once within $k$ runs~\cite{yue_2025_rlvr}.

\begin{figure}[t]
\centering
\begin{tikzpicture}
\begin{groupplot}[group style={group size=2 by 1, horizontal sep=1.15cm}, scale only axis, width=2.75cm, height=2.5cm, axis lines=box, axis line style={black, line width=0.5pt}, tick style={black, line width=0.4pt}, major tick length=1.6pt, tick align=outside, xtick pos=bottom, ytick pos=left, xmajorgrids, ymajorgrids, major grid style={black!40, dashed, line width=0.25pt}, tick label style={font=\scriptsize}, label style={font=\scriptsize}, title style={font=\footnotesize, yshift=-3pt, text height=1.6ex, text depth=0.4ex}]
\nextgroupplot[xmode=log, log basis x=2, xmin=0.85, xmax=18.5, xtick={1,2,4,8,16}, xticklabels={1,2,4,8,16}, xlabel={$k$}, ymin=20, ymax=85, ytick={20,40,60,80}, title={(a) pass@$k$, \%}, legend to name=legpassk, legend columns=-1, legend cell align=left, legend style={font=\scriptsize, draw=none, /tikz/every even column/.append style={column sep=10pt}}]
\addplot[cOrig, line width=1.4pt, mark=triangle*, mark size=2.2pt, mark options={solid, fill=cOrig, draw=white, line width=0.25pt}] coordinates {(1,39.73) (2,52.87) (3,59.28) (4,63.19) (5,65.86) (6,67.83) (7,69.32) (8,70.51) (9,71.46) (10,72.26) (11,72.94) (12,73.52) (13,74.03) (14,74.49) (15,74.89) (16,75.26)};
\addplot[cRew, line width=1.4pt, mark=triangle*, mark size=2.2pt, mark options={solid, fill=cRew, draw=white, line width=0.25pt}] coordinates {(1,26.72) (2,36.59) (3,42.17) (4,45.96) (5,48.8) (6,51.05) (7,52.89) (8,54.46) (9,55.84) (10,57.08) (11,58.22) (12,59.3) (13,60.33) (14,61.32) (15,62.28) (16,63.22)};
\addlegendentry{Original skill}\addlegendentry{Round-30 skill}
\nextgroupplot[xmode=log, log basis x=2, xmin=0.85, xmax=18.5, xtick={1,2,4,8,16}, xticklabels={1,2,4,8,16}, xlabel={$k$}, ymin=-25, ymax=2, ytick={-20,-15,-10,-5,0}, title={(b) Round 30 $-$ original, pp}]
\addplot[fill=cRew!22, draw=none, forget plot] coordinates {(1,-19.56) (2,-23.7) (3,-24.71) (4,-24.4) (5,-24.47) (6,-24.15) (7,-23.74) (8,-23.76) (9,-23.1) (10,-23.12) (11,-22.64) (12,-22.15) (13,-21.76) (14,-21.04) (15,-21.18) (16,-20.43) (16,-4.14) (15,-4.75) (14,-5.49) (13,-6.27) (12,-6.64) (11,-7.18) (10,-7.76) (9,-8.33) (8,-8.76) (7,-9.14) (6,-9.27) (5,-10.19) (4,-10.37) (3,-10.07) (2,-9.32) (1,-6.96)} -- cycle;
\addplot[black!55, dashed, line width=0.8pt, domain=0.85:18.5, samples=2, forget plot] {0};
\addplot[cRew, line width=1.4pt, mark=triangle*, mark size=2.2pt, mark options={solid, fill=cRew, draw=white, line width=0.25pt}, forget plot] coordinates {(1,-13.0) (2,-16.28) (3,-17.12) (4,-17.22) (5,-17.06) (6,-16.78) (7,-16.43) (8,-16.04) (9,-15.63) (10,-15.19) (11,-14.72) (12,-14.22) (13,-13.71) (14,-13.17) (15,-12.61) (16,-12.05)};
\end{groupplot}
\node[anchor=south, yshift=1pt] at (current bounding box.north) {\pgfplotslegendfromname{legpassk}};
\end{tikzpicture}
\caption{\textbf{With up to sixteen attempts per task, the round-30 skill stays below the original skill.} (a) pass@$k$ of \modelLuna on SkillsBench, the chance that at least one of $k$ runs passes a task. (b) Round-30 skill minus original skill, with its 95\% interval over tasks.}
\label{fig:suppl_passk}
\end{figure}
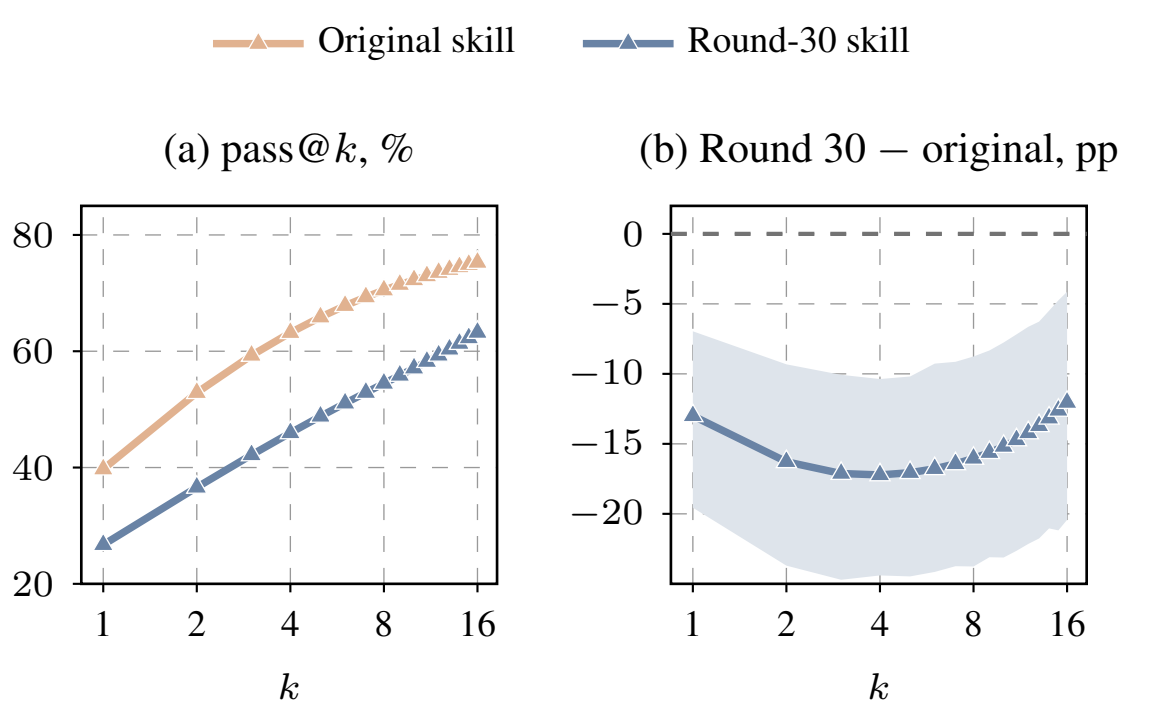

\paragraph{The round-30 skill lies below the original skill at every $k$.}
Even with sixteen attempts per task, the round-30 skill remains well below the original: pass@16 reaches 75.3\% with the original skill but only 63.2\% with the rewritten skill (\cref{fig:suppl_passk}).
14 tasks pass at least once with the original skill but never in sixteen runs with the rewritten skill; only 2 tasks show the reverse.
The round-30 skill stays below the original at every $k$ from 1 to 16, and the 95\% interval of the difference lies below zero at every $k$.
Rewriting thus reduces both the chance of success in a single run and the set of tasks the agent solves within sixteen runs.

\subsection{Repeated Evaluations}
\label{sec:suppl_repeats}

The same agent with the same skill can pass a task in one run and fail it in the next, so a single evaluation leaves room for chance.
To check that the decline and the recovery under restoring do not come from this chance, we use the repeated runs of \modelLuna and \modelSol on SkillsBench.
With the original and the round-30 skill, \modelLuna runs every task at least sixteen times and \modelSol at least four times, and with the skills of rounds 1, 5 and 10 and the three restored versions, every task runs at least twice.

\begin{figure*}[t]
\centering
\begin{tikzpicture}
\begin{groupplot}[group style={group size=3 by 2, horizontal sep=1.2cm, vertical sep=0.45cm}, scale only axis, width=3.6cm, height=2.2cm, axis lines=box, axis line style={black, line width=0.5pt}, tick style={black, line width=0.4pt}, major tick length=1.6pt, tick align=outside, xtick pos=bottom, ytick pos=left, xmajorgrids, ymajorgrids, major grid style={black!40, dashed, line width=0.25pt}, tick label style={font=\scriptsize}, label style={font=\scriptsize}, title style={font=\footnotesize, yshift=-3pt, text height=1.6ex, text depth=0.4ex}]
\nextgroupplot[xmin=-2, xmax=32, xtick={0,10,20,30}, ymin=-25, ymax=4, ytick={-20,-10,0}, title={(a) Change from original, pp}, ylabel={{\footnotesize \modelLuna}}, xticklabels={}, legend to name=legrep, legend columns=3, legend cell align=left, legend style={font=\scriptsize, draw=none, /tikz/every even column/.append style={column sep=10pt}}]
\addplot[fill=cRew!22, draw=none, forget plot] coordinates {(0,0) (1,-14.01) (5,-21.5) (10,-20.53) (30,-19.56) (30,-6.96) (10,-7.38) (5,-6.76) (1,-2.77) (0,0)} -- cycle;
\addplot[cOrig, line width=1.4pt, domain=-2:32, samples=2, forget plot] {0};
\addplot[cRew, line width=1.4pt, mark=triangle*, mark size=2.2pt, mark options={solid, fill=cRew, draw=white, line width=0.25pt}, forget plot] coordinates {(0,0) (1,-8.36) (5,-14.06) (10,-13.87) (30,-13.0)};
\addlegendimage{cOrig, line width=1.4pt}\addlegendentry{Original skill}
\addlegendimage{cRew, line width=1.4pt, mark=triangle*, mark size=2.2pt, mark options={solid, fill=cRew, draw=white, line width=0.25pt}}\addlegendentry{Rewritten skill, mean over runs}
\addlegendimage{only marks, mark=*, mark size=1.3pt, black}\addlegendentry{One full evaluation}
\addlegendimage{area legend, fill=cNone, draw=none}\addlegendentry{0/10 items restored}
\addlegendimage{area legend, fill=cSome, draw=none}\addlegendentry{$\geq$1/10 items restored}
\addlegendimage{area legend, fill=cAll, draw=none}\addlegendentry{All lost items restored}
\nextgroupplot[xmin=-0.35, xmax=1.35, xtick={0,1}, xticklabels={Original,Round 30}, xmajorgrids=false, ymin=15, ymax=60, ytick={20,30,40,50,60}, title={(b) Full evaluations, pass rate, \%}]
\addplot[black!30, line width=0.4pt, forget plot] coordinates {(0,45.98) (1,20.69)};
\addplot[black!30, line width=0.4pt, forget plot] coordinates {(0,35.63) (1,20.69)};
\addplot[black!30, line width=0.4pt, forget plot] coordinates {(0,36.78) (1,32.18)};
\addplot[black!30, line width=0.4pt, forget plot] coordinates {(0,44.83) (1,27.59)};
\addplot[black!30, line width=0.4pt, forget plot] coordinates {(0,39.08) (1,33.33)};
\addplot[black!30, line width=0.4pt, forget plot] coordinates {(0,39.08) (1,33.33)};
\addplot[black!30, line width=0.4pt, forget plot] coordinates {(0,43.68) (1,28.74)};
\addplot[black!30, line width=0.4pt, forget plot] coordinates {(0,39.08) (1,25.29)};
\addplot[black!30, line width=0.4pt, forget plot] coordinates {(0,40.23) (1,25.29)};
\addplot[black!30, line width=0.4pt, forget plot] coordinates {(0,40.23) (1,33.33)};
\addplot[black!30, line width=0.4pt, forget plot] coordinates {(0,39.08) (1,24.14)};
\addplot[black!30, line width=0.4pt, forget plot] coordinates {(0,43.68) (1,21.84)};
\addplot[black!30, line width=0.4pt, forget plot] coordinates {(0,39.08) (1,20.69)};
\addplot[black!30, line width=0.4pt, forget plot] coordinates {(0,29.89) (1,26.44)};
\addplot[black!30, line width=0.4pt, forget plot] coordinates {(0,40.23) (1,25.29)};
\addplot[black!30, line width=0.4pt, forget plot] coordinates {(0,42.53) (1,28.74)};
\addplot[only marks, mark=triangle*, mark size=2.2pt, mark options={solid, fill=cOrig, draw=white, line width=0.25pt}, forget plot] coordinates {(0,45.98) (0,35.63) (0,36.78) (0,44.83) (0,39.08) (0,39.08) (0,43.68) (0,39.08) (0,40.23) (0,40.23) (0,39.08) (0,43.68) (0,39.08) (0,29.89) (0,40.23) (0,42.53)};
\addplot[only marks, mark=triangle*, mark size=2.2pt, mark options={solid, fill=cRew, draw=white, line width=0.25pt}, forget plot] coordinates {(1,20.69) (1,20.69) (1,32.18) (1,27.59) (1,33.33) (1,33.33) (1,28.74) (1,25.29) (1,25.29) (1,33.33) (1,24.14) (1,21.84) (1,20.69) (1,26.44) (1,25.29) (1,28.74)};
\nextgroupplot[xmin=-0.5, xmax=3.5, xtick={0,1,2,3}, xticklabels={R30,0/10,$\geq$1/10,All}, xmajorgrids=false, ymin=0, ymax=60, ytick={0,20,40,60}, title={(c) Restoration, pass rate, \%}]
\addplot[cOrig, line width=1.4pt, domain=-0.5:3.5, samples=2, forget plot] {39.73};
\addplot[ybar, bar width=9pt, fill=cRew, draw=none, forget plot] coordinates {(0,26.72)};
\addplot[ybar, bar width=9pt, fill=cNone, draw=none, forget plot] coordinates {(1,37.16)};
\addplot[ybar, bar width=9pt, fill=cSome, draw=none, forget plot] coordinates {(2,29.89)};
\addplot[ybar, bar width=9pt, fill=cAll, draw=none, forget plot] coordinates {(3,33.33)};
\addplot[only marks, mark=*, mark size=1.3pt, black, forget plot] coordinates {(0,20.69) (0,20.69) (1,35.63) (1,35.63) (2,26.44) (2,27.59) (3,34.48) (3,33.33)};
\nextgroupplot[xmin=-2, xmax=32, xtick={0,10,20,30}, ymin=-25, ymax=4, ytick={-20,-10,0}, ylabel={{\footnotesize \modelSol}}, xlabel={Round}]
\addplot[fill=cRew!22, draw=none, forget plot] coordinates {(0,0) (1,-12.34) (5,-16.56) (10,-21.21) (30,-17.18) (30,-6.49) (10,-8.48) (5,-4.11) (1,-2.85) (0,0)} -- cycle;
\addplot[cOrig, line width=1.4pt, domain=-2:32, samples=2, forget plot] {0};
\addplot[cRew, line width=1.4pt, mark=triangle*, mark size=2.2pt, mark options={solid, fill=cRew, draw=white, line width=0.25pt}, forget plot] coordinates {(0,0) (1,-7.31) (5,-9.99) (10,-14.59) (30,-11.52)};
\nextgroupplot[xmin=-0.35, xmax=1.35, xtick={0,1}, xticklabels={Original,Round 30}, xmajorgrids=false, ymin=15, ymax=60, ytick={20,30,40,50,60}]
\addplot[black!30, line width=0.4pt, forget plot] coordinates {(0,55.17) (1,41.38)};
\addplot[black!30, line width=0.4pt, forget plot] coordinates {(0,50.57) (1,39.08)};
\addplot[black!30, line width=0.4pt, forget plot] coordinates {(0,55.17) (1,37.93)};
\addplot[black!30, line width=0.4pt, forget plot] coordinates {(0,50.57) (1,42.53)};
\addplot[only marks, mark=triangle*, mark size=2.2pt, mark options={solid, fill=cOrig, draw=white, line width=0.25pt}, forget plot] coordinates {(0,55.17) (0,50.57) (0,55.17) (0,50.57)};
\addplot[only marks, mark=triangle*, mark size=2.2pt, mark options={solid, fill=cRew, draw=white, line width=0.25pt}, forget plot] coordinates {(1,41.38) (1,39.08) (1,37.93) (1,42.53)};
\nextgroupplot[xmin=-0.5, xmax=3.5, xtick={0,1,2,3}, xticklabels={R30,0/10,$\geq$1/10,All}, xmajorgrids=false, ymin=0, ymax=60, ytick={0,20,40,60}, xlabel={Lost items restored}]
\addplot[cOrig, line width=1.4pt, domain=-0.5:3.5, samples=2, forget plot] {52.33};
\addplot[ybar, bar width=9pt, fill=cRew, draw=none, forget plot] coordinates {(0,40.8)};
\addplot[ybar, bar width=9pt, fill=cNone, draw=none, forget plot] coordinates {(1,48.66)};
\addplot[ybar, bar width=9pt, fill=cSome, draw=none, forget plot] coordinates {(2,43.68)};
\addplot[ybar, bar width=9pt, fill=cAll, draw=none, forget plot] coordinates {(3,47.89)};
\addplot[only marks, mark=*, mark size=1.3pt, black, forget plot] coordinates {(0,41.38) (0,39.08) (1,48.28) (1,49.43) (2,44.83) (2,44.83) (3,45.98) (3,48.28)};
\end{groupplot}
\node[anchor=south, yshift=1pt] at (current bounding box.north) {\pgfplotslegendfromname{legrep}};
\end{tikzpicture}
\caption{\textbf{Over repeated runs, the rewritten skill stays below the original skill, and putting back the items outside the prior recovers the most.} (a) Change in pass rate on SkillsBench from the original skill, with every run of a task averaged, and its 95\% interval over tasks. (b) Full evaluations of the benchmark, each taking the $j$-th run of every task; a line joins the original and the round-30 skill of the same evaluation. (c) Pass rate of the round-30 skill and of the round-30 skill with the lost 0/10 items outside the prior, the other lost items in the $\geq$1/10 group or all lost items restored. Bars show averages over runs; dots show each full evaluation.}
\label{fig:suppl_repeats}
\end{figure*}
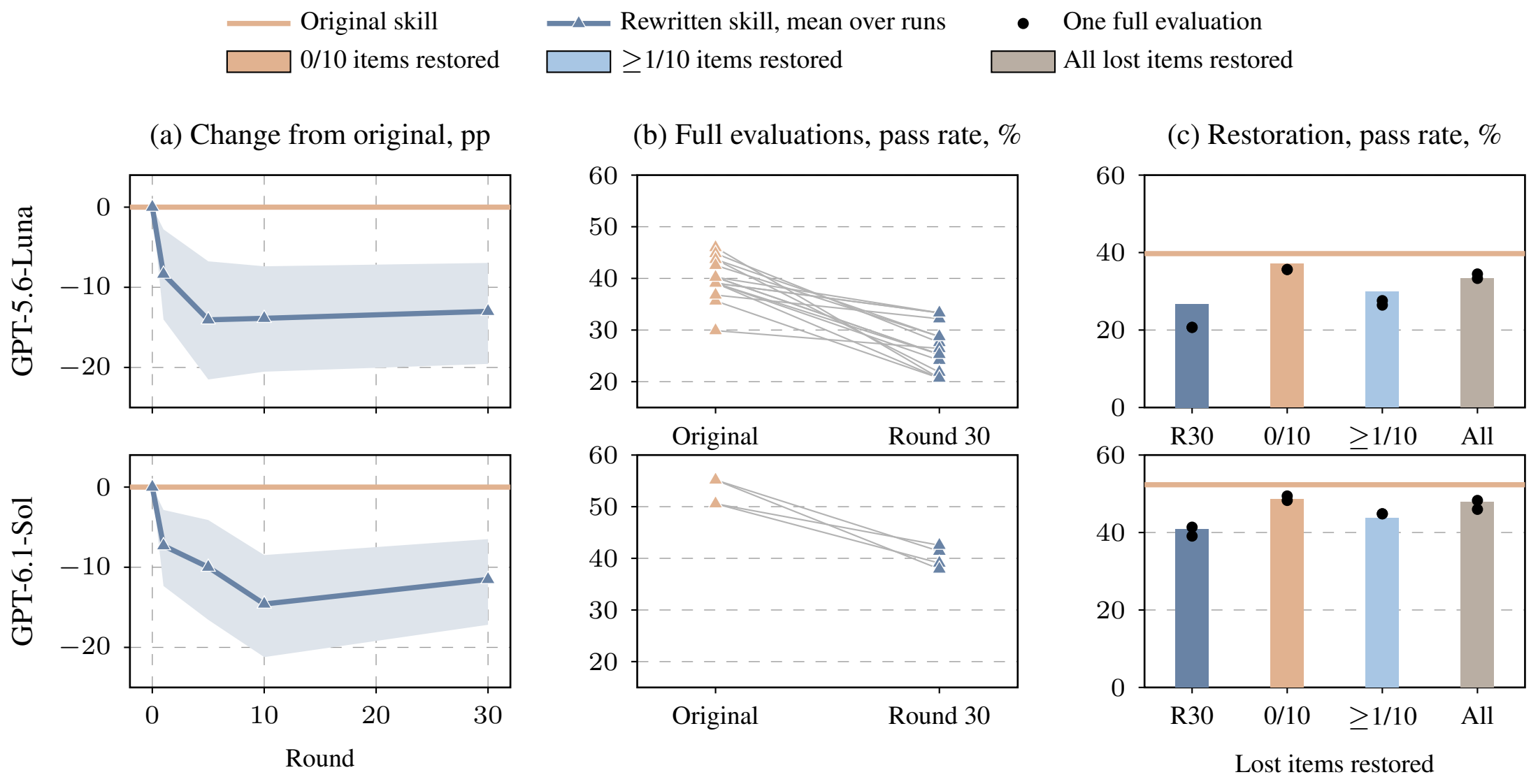

\paragraph{Every full evaluation shows the decline.}
The round-30 skill falls below the original in 16 of 16 full evaluations for \modelLuna and 4 of 4 for \modelSol, with mean losses of 13.2 and 12.6 points, respectively (\cref{fig:suppl_repeats}b).
Each full evaluation takes the $j$-th run of every task.
For \modelLuna, mean pass rates with the original and round-30 skills are 39.9\% and 26.7\%, with standard deviations of 3.9 and 4.6 points.
For \modelSol, the mean pass rates are 52.9\% and 40.2\%.

\paragraph{Averaged over runs, the decline follows the main text.}
With every run of a task averaged, the pass rate of \modelLuna lies 8.4 points below the original skill after one round and 13.0 after thirty, and that of \modelSol{} 7.3 and 11.5 points, and the 95\% interval of the change lies below zero after every round (\cref{fig:suppl_repeats}a).

\paragraph{Putting back the items outside the prior recovers the most.}
Averaged over runs, putting the lost items outside the prior back into the round-30 skill raises the pass rate by 10.4 points for \modelLuna and 7.9 points for \modelSol. Putting back the other lost items raises it by 3.2 and 2.9 points (\cref{fig:suppl_repeats}c).
In each of the four full evaluations of the restored versions, the items outside the prior recover more than the other items, by 6.3 points on average.
The decline and the recovery are thus not accidents of single runs.

\subsection{Repeated Rewriting Chains}
\label{sec:suppl_chains}

Rewriting samples from the model, so the path of one rewriting chain may differ from that of another.
On LogicBench and MedCalc-Bench, \modelLuna, \modelSol, \modelMimo, \modelQwen and \modelQwenFlash therefore rewrite every skill in two further independent chains of thirty rounds, and on $\tau^2$-bench every policy in ten chains in all.
The further chains share the self-written skills of the first chain and are measured and judged as in the main text.

\input{fig/chains}
\begin{table}[t]
\centering\footnotesize
\setlength{\tabcolsep}{4pt}
\fitcols{2}{Kept after round 30, \%}{97--100}
\begin{tabular}{@{}lWW>{\columncolor{hlCol}}W@{}}
\toprule
& \multicolumn{2}{c}{Kept after round 30, \%} & \multicolumn{1}{c}{}\\
\cmidrule(lr){2-3}
Model & 0/10 & 10/10 & $\Delta$, pp\\
\midrule
\multicolumn{4}{@{}l}{\textit{LogicBench, three chains per skill}}\\
\modelLuna & 18--21 & 77--83 & 59--62\\
\modelSol & 11--11 & 80--83 & 69--72\\
\modelMimo & 25--28 & 85--93 & 60--65\\
\modelQwen & 18--26 & 88--90 & 62--72\\
\modelQwenFlash & 15--17 & 73--77 & 56--59\\
\midrule
\multicolumn{4}{@{}l}{\textit{MedCalc-Bench, three chains per skill}}\\
\modelLuna & 19--22 & 83--87 & 64--67\\
\modelSol & 14--18 & 84--86 & 66--72\\
\modelMimo & 19--22 & 88--92 & 69--70\\
\modelQwen & 36--41 & 83--89 & 44--53\\
\modelQwenFlash & 4--6 & 73--78 & 66--75\\
\midrule
\multicolumn{4}{@{}l}{\textit{$\tau^2$-bench, ten chains per policy}}\\
\modelLuna & 9--51 & 72--90 & 28--74\\
\modelSol & 58--67 & 77--97 & 15--37\\
\modelMimo & 54--69 & 97--100 & 31--46\\
\modelQwen & 49--67 & 91--100 & 32--42\\
\modelQwenFlash & 3--47 & 69--100 & 47--90\\
\bottomrule
\end{tabular}
\caption{\textbf{Every rewriting chain keeps the items that all self-written skills state far more often than the items outside the prior.} Range over the independent rewriting chains of each model of the share of items kept after thirty rounds, for the 0/10 items outside the prior and for the 10/10 items that all ten self-written skills or policies state, and of their difference $\Delta$ in pp.}
\label{tab:suppl_chains}
\end{table}

\paragraph{The distance follows nearly the same path in every chain.}
In every chain, the skills move toward the self-written skills over the first rounds and then level off (\cref{fig:suppl_chains}), and after thirty rounds the distances of the three chains of a model differ by only 0.02 units on average.

\paragraph{Every chain loses the items outside the prior first.}
On LogicBench and MedCalc-Bench, the share of the items outside the prior that a model keeps differs by 3 points on average between its chains, and in every chain the items that all self-written skills state are kept more often, by 64 points on average (\cref{tab:suppl_chains}).
On $\tau^2$-bench, with two policies and few items per group, single chains vary more, yet in each of the 50 chains the items outside the prior are kept less often, by 45 points on average.
The movement toward the prior and the selective loss thus do not hinge on a single rewriting chain.

\subsection{Deleting Content from the Original Skill}
\label{sec:suppl_delete}

Restoring puts content back into a rewritten skill; the converse test takes content out of the original skill.
If the items outside the prior carry more of the value of a skill, deleting them should lower the pass rate more than deleting as much of the other content.
From each original SkillsBench skill, we therefore delete about a quarter of the body in two ways, either only items outside the prior or only items that some self-written skill states, with the same number of words in both versions of a skill.
Where one kind of item makes up less than a quarter of the body, both versions delete as many words as that kind holds, so that on average the two versions delete 23.4\% and 23.1\% of the body for \modelLuna and 22.5\% and 22.2\% for \modelSol.
Each model separates the two kinds of item with its own self-written skills.
The deletion removes whole items and rewrites no text, keeping the front matter, removing a heading only when everything under it goes, and moving nested list entries up.
\modelLuna and \modelSol run both versions once per task, in the same batch as the original skill and under the protocol of the main text.

\begin{table}[t]
\centering\footnotesize
\setlength{\tabcolsep}{3pt}
\resizebox{\columnwidth}{!}{\begin{tabular}{@{}lccc@{}}
\toprule
$\Delta$ pass rate, pp & Both models & \modelLuna & \modelSol\\
\midrule
\multicolumn{4}{@{}l}{\textit{Against the original skill}}\\
0/10 items deleted & $-4.6$ & $-5.7$ & $-3.4$\\
$\geq$1/10 items deleted & $+0.6$ & $+2.3$ & $-1.1$\\
\midrule
\multicolumn{4}{@{}l}{\textit{Between the two deletions}}\\
\rowcolor{hlRow}0/10 $-$ $\geq$1/10 & $-5.2$ & $-8.0$ & $-2.3$\\
\bottomrule
\end{tabular}}
\caption{\textbf{Deleting the items outside the prior lowers the pass rate more than deleting as many words of the other items.} Change in pass rate on SkillsBench when the same number of words is deleted from the original skill, only from the 0/10 items outside the prior or only from the other items in the $\geq$1/10 group. Each version is run in the same batch as the original skill.}
\label{tab:suppl_deleq}
\end{table}

\paragraph{Deleting the items outside the prior costs more.}
With similar amounts of text removed, the two deletions lead to different changes in performance.
On average over both models, deleting items outside the prior lowers the pass rate by 4.6 points, while deleting the other items changes it by +0.6 points, a mean difference of 5.2 points (\cref{tab:suppl_deleq}).
Both models point the same way, as the deletion outside the prior costs \modelLuna{} 8.0 points and \modelSol{} 2.3 points more than the other deletion.
Taking content out of the original skill thus mirrors putting it back into the rewritten skill: similar amounts of text matter more when the content lies outside the prior.

\subsection{Rewritten Skills Run by the Other Model}
\label{sec:suppl_docexec}

Whether the loss depends on who rewrites a skill and who runs it can be tested by swapping the two.
We give the round-30 skills of \modelLuna and of \modelSol to both models as agents on SkillsBench, once per task, in the batch of the deletion experiment and against the original skill of the same batch.

\begin{table}[t]
\centering\footnotesize
\setlength{\tabcolsep}{3pt}
\resizebox{\columnwidth}{!}{\begin{tabular}{@{}lcccc@{}}
\toprule
& & \multicolumn{2}{c}{Round-30 skill rewritten by} & $\Delta$ own $-$\\
\cmidrule(lr){3-4}
Agent & Original & \modelLuna & \modelSol & other's, pp\\
\midrule
\modelLuna & 36.8 & \cellcolor{hlAuth}31.0 & 27.6 & $+3.4$\\
\modelSol & 44.8 & 42.5 & \cellcolor{hlAuth}37.9 & $-4.6$\\
\bottomrule
\end{tabular}}
\caption{\textbf{Both agents do worse with either round-30 skill, and about as badly with their own rewrite as with the other one.} Pass rate on SkillsBench, in percent, with the original skill and with the round-30 skills rewritten by each model, all run in the same batch, and the difference between the own and the other rewrite, in pp. Shaded cells: an agent running its own rewrite.}
\label{tab:suppl_docexec}
\end{table}

\paragraph{Both agents lose performance with either rewrite.}
Whether an agent uses its own rewrite or the other model's, all four combinations fall below the original skill of the same batch, by 6.0 points on average (\cref{tab:suppl_docexec}).
Swapping the executor between these two models does not eliminate the loss.
This agrees with the overlap of the items outside the prior across models (\cref{sec:suppl_overlap}), since what the rewriting of one model gives up is largely what the other model does not write on its own either.

\FloatBarrier
\section{Implementation Details}
\label{sec:suppl_impl}

\subsection{Evaluation Protocol}
\label{sec:suppl_eval}

\paragraph{Tasks and harnesses.}
Every run takes place in the task's container under benchflow, with the skill placed in the directory from which the harness loads skills.
\modelLuna runs in Codex through the codex-acp adapter 1.6.2 and \modelSol through codex-acp 2.1.1.
\modelMimo, \modelQwen and \modelQwenFlash run in OpenHands, and \modelHaiku runs in Claude Code through claude-agent-acp 0.40.0 at its default effort.
On LogicBench and MedCalc-Bench, every model answers at its lowest reasoning setting, minimal for the GPT models, without thinking for \modelMimo{} and the Qwen models, and at low effort for \modelHaiku.

\paragraph{Rewriting and judging.}
\modelLuna and \modelSol rewrite the skills through Codex at high reasoning effort with its default tools, which include web search.
\modelMimo, \modelQwen and \modelQwenFlash rewrite in a single API call without a system prompt or tools at their default thinking setting, and \modelHaiku rewrites through Claude Code with every tool turned off and an empty system prompt.
Every model writes its self-written skills under the same settings as its rewrites, and \modelJudge judges the items and locates the passages for restoring through Codex.

\paragraph{Pass rates.}
The pass rate of a condition is the share of the 87 tasks that its runs pass, and a difference between two conditions compares them on the same tasks.

\subsection{Items and Judging}
\label{sec:suppl_items}

\paragraph{Items.}
Every non-blank line of an original skill belongs to exactly one item, and the front matter forms one more item.
A heading is one heading line, a list entry is a line with a list marker together with the lines that continue it, a table is a run of table lines, a code block is a fenced block with its fences, and a paragraph is a run of non-blank lines that starts no other kind of item.
The 87 skills contain 6,605 items with 83,950 words, namely 3,085 list entries, 1,389 headings, 1,292 paragraphs, 684 code blocks, 87 front matter items and 68 tables.

\paragraph{Judging.}
\modelJudge judges all items of a skill in a single call against one candidate document, which is either a self-written skill or a rewritten version.
Against a self-written skill, an item counts as stated only when the document conveys the same content, and the same nouns, file names or numbers without the same content do not count.
Against a rewritten skill, an item counts as kept when it is present verbatim or reworded with the same content, including code that does the same thing, and as lost when it is weakened to a partial, optional or hedged form, dropped, or changed to different content on the same point.
The judge is told that a change may be harmless or even an improvement, so that the labels describe what happened to an item without presuming damage.
An item counts as stated by a self-written skill when the label is \texttt{SAME}.
It counts as kept by a rewrite when the label is \texttt{KEPT} or \texttt{REWORDED} and as lost when it is \texttt{WEAKENED}, \texttt{DROPPED} or \texttt{CHANGED}, and items labeled \texttt{UNCLEAR} are left out.

\subsection{Restoration Construction}
\label{sec:suppl_restore}

\paragraph{Common base.}
For every item lost by round 30, \modelJudge locates the passage of the round-30 skill that carries its weakened or changed version, and this passage is removed, or within a long passage only the sentences that carry lost items.
The result is the common base of the three restored versions.

\paragraph{Putting items back.}
Each restored version puts the original text of its lost items back into the common base in their original order, where the removed passage stood or after the line the locator names, and at the end where no place is found.
Checks confirm that every item of the condition returns, that no item of the other kind does, and that the base is identical across the three versions.

\subsection{Cost of \method}
\label{sec:suppl_cost}

\method spends its cost once per skill and model, before any run of the agent.
One call writes the topic, the model writes ten self-written skills from it, and the judge compares the items of the original skill with each of them, so a skill takes one topic call, ten generation calls and ten judge calls.
On SkillsBench with \modelLuna, a self-written skill holds 1,031 words on average against 965 words of the original skill, so the model writes about ten times the length of the skill, and each judge call reads the skill and one self-written skill.
Marking itself is a text operation that adds the note and 31 tags on average, 129 words in all, and the agent then runs as it does with the original skill.

\FloatBarrier

\section{Future Directions}
\label{sec:suppl_future}

One direction follows directly from these results.
A rewrite that leaves the items outside the prior untouched and changes only the rest could keep what feedback-free self-evolution now loses, since restoring these items recovers much of the lost pass rate, and rewrites that preserve the operational anchors of a skill already keep its quality under compression~\cite{xing_2026_skillremember}.
The self-written skills locate these items without any run of the agent, so such a guard needs no feedback.
How much feedback a self-evolution loop needs to protect this knowledge on its own remains an open question.

\section{Limitations}
\label{sec:suppl_limitations}

\paragraph{Granularity of skill updates.}
We rewrite only the skill document and leave the accompanying scripts and reference files unchanged.
Each SkillsBench task uses one preselected skill.
We have not studied the retrieval or composition of multiple skills, or the joint evolution of skill documents and code.


{
\small
\begin{thebibliography}{59}
\providecommand{\natexlab}[1]{#1}

\bibitem[Acerbi and Stubbersfield(2023)]{acerbi_2023_transmission}
Alberto Acerbi and Joseph~M. Stubbersfield.
\newblock Large language models show human-like content biases in transmission
  chain experiments.
\newblock \emph{Proceedings of the National Academy of Sciences}, 120\penalty0
  (44):\penalty0 e2313790120, 2023.

\bibitem[Agrawal et~al.(2026)Agrawal, Tan, Soylu, Ziems, Khare, Opsahl-Ong,
  Singhvi, Shandilya, Ryan, Jiang, Potts, Sen, Dimakis, Stoica, Klein, Zaharia,
  and Khattab]{agrawal_2025_gepa}
Lakshya~A Agrawal, Shangyin Tan, Dilara Soylu, Noah Ziems, Rishi Khare, Krista
  Opsahl-Ong, Arnav Singhvi, Herumb Shandilya, Michael~J Ryan, Meng Jiang,
  Christopher Potts, Koushik Sen, Alex Dimakis, Ion Stoica, Dan Klein, Matei
  Zaharia, and Omar Khattab.
\newblock {GEPA}: Reflective prompt evolution can outperform reinforcement
  learning.
\newblock In \emph{International Conference on Learning Representations}, 2026.

\bibitem[{Anthropic}(2026)]{anthropic_2026_haiku55}
{Anthropic}.
\newblock System card: {Claude Haiku 5.5}.
\newblock \url{https://www.anthropic.com/claude-haiku-5-5-system-card}, 2026.
\newblock Published October 7, 2026.

\bibitem[Barres et~al.(2026)Barres, Dong, Ray, Si, and
  Narasimhan]{barres_2025_tau2}
Victor Barres, Honghua Dong, Soham Ray, Xujie Si, and Karthik~R Narasimhan.
\newblock $\tau^2$-bench: Evaluating conversational agents in a dual-control
  environment.
\newblock In \emph{International Conference on Machine Learning}, 2026.

\bibitem[Chen et~al.(2026)Chen, Wei, Yao, Lu, Qian, Wu, and
  He]{chen_2026_judgegates}
Chenle Chen, Yangbo Wei, Chao Yao, Shaoqiang Lu, Junhong Qian, Chen Wu, and Lei
  He.
\newblock Competence, not accuracy: A diagnostic for reference-free judge gates
  in skill optimization.
\newblock arXiv preprint arXiv:2608.18719, 2026.

\bibitem[Chen et~al.(2021)Chen, Tworek, Jun, Yuan, de~Oliveira~Pinto, Kaplan,
  Edwards, Burda, Joseph, Brockman, Ray, Puri, Krueger, Petrov, Khlaaf, Sastry,
  Mishkin, Chan, Gray, Ryder, Pavlov, Power, Kaiser, Bavarian, Winter, Tillet,
  Such, Cummings, Plappert, Chantzis, Barnes, Herbert-Voss, Guss, Nichol,
  Paino, Tezak, Tang, Babuschkin, Balaji, Jain, Saunders, Hesse, Carr, Leike,
  Achiam, Misra, Morikawa, Radford, Knight, Brundage, Murati, Mayer, Welinder,
  McGrew, Amodei, McCandlish, Sutskever, and Zaremba]{chen_2021_codex}
Mark Chen, Jerry Tworek, Heewoo Jun, Qiming Yuan, Henrique~Ponde de
  Oliveira~Pinto, Jared Kaplan, Harri Edwards, Yuri Burda, Nicholas Joseph,
  Greg Brockman, Alex Ray, Raul Puri, Gretchen Krueger, Michael Petrov, Heidy
  Khlaaf, Girish Sastry, Pamela Mishkin, Brooke Chan, Scott Gray, Nick Ryder,
  Mikhail Pavlov, Alethea Power, Lukasz Kaiser, Mohammad Bavarian, Clemens
  Winter, Philippe Tillet, Felipe~Petroski Such, Dave Cummings, Matthias
  Plappert, Fotios Chantzis, Elizabeth Barnes, Ariel Herbert-Voss,
  William~Hebgen Guss, Alex Nichol, Alex Paino, Nikolas Tezak, Jie Tang, Igor
  Babuschkin, Suchir Balaji, Shantanu Jain, William Saunders, Christopher
  Hesse, Andrew~N. Carr, Jan Leike, Josh Achiam, Vedant Misra, Evan Morikawa,
  Alec Radford, Matthew Knight, Miles Brundage, Mira Murati, Katie Mayer, Peter
  Welinder, Bob McGrew, Dario Amodei, Sam McCandlish, Ilya Sutskever, and
  Wojciech Zaremba.
\newblock Evaluating large language models trained on code.
\newblock arXiv preprint arXiv:2107.03374, 2021.

\bibitem[Chen et~al.(2024)Chen, Li, Yang, Yu, Lin, and
  He]{chen_2024_automanual}
Minghao Chen, Yihang Li, Yanting Yang, Shiyu Yu, Binbin Lin, and Xiaofei He.
\newblock {AutoManual}: Constructing instruction manuals by {LLM} agents via
  interactive environmental learning.
\newblock In \emph{Advances in Neural Information Processing Systems}, 2024.

\bibitem[Destefanis et~al.(2026)Destefanis, Graziotin, Vaccargiu, and
  Ortu]{destefanis_2026_gitskills}
Giuseppe Destefanis, Daniel Graziotin, Matteo Vaccargiu, and Marco Ortu.
\newblock {GitSkills}: A dataset of agent skills on {GitHub}.
\newblock arXiv preprint arXiv:2608.10906, 2026.

\bibitem[Duan et~al.(2026)Duan, Shi, Mao, Zhang, Liang, Piao, Wu, Yao, Huang,
  Huang, Sun, Yan, and Zhang]{duan_2026_repo2skillevo}
Chenyuan Duan, Ge Shi, Zineng Mao, Ge Zhang, Hao Liang, Yinzhu Piao, Yuchen Wu,
  Zhixin Yao, Kaiyu Huang, Wenhao Huang, Linzhuang Sun, Shen Yan, and Wentao
  Zhang.
\newblock {Repo2Skill-Evo}: Repository skills go stale in silence.
\newblock arXiv preprint arXiv:2608.21964, 2026.

\bibitem[Gao et~al.(2026{\natexlab{a}})Gao, Chen, Wang, Guo, Pang, Liu, Shen,
  and Cheng]{gao_2026_skillaudit}
Haowen Gao, Haoran Chen, Can Wang, Shasha Guo, Liang Pang, Zhaoyang Liu, Huawei
  Shen, and Xueqi Cheng.
\newblock {SelfEvoSkill}: Paired execution audits for skill revision without
  external outcome supervision.
\newblock arXiv preprint arXiv:2606.14239, 2026{\natexlab{a}}.

\bibitem[Gao et~al.(2026{\natexlab{b}})Gao, Geng, Hua, Hu, Juan, Liu, Liu, Qiu,
  Qi, Ren, Wu, Wang, Xiao, Zhou, Zhang, Zhang, Xiang, Fang, Zhao, Liu, Qian,
  Wang, Hu, Wang, Wu, Ji, and Wang]{gao_2025_selfevolving}
{Huan-ang} Gao, Jiayi Geng, Wenyue Hua, Mengkang Hu, Xinzhe Juan, Hongzhang
  Liu, Shilong Liu, Jiahao Qiu, Xuan Qi, Qihan Ren, Yiran Wu, Hongru Wang, Han
  Xiao, Yuhang Zhou, Shaokun Zhang, Jiayi Zhang, Jinyu Xiang, Yixiong Fang,
  Qiwen Zhao, Dongrui Liu, Cheng Qian, Zhenhailong Wang, Minda Hu, Huazheng
  Wang, Qingyun Wu, Heng Ji, and Mengdi Wang.
\newblock A survey of self-evolving agents: What, when, how, and where to
  evolve on the path to artificial super intelligence.
\newblock \emph{Transactions on Machine Learning Research}, 2026{\natexlab{b}}.

\bibitem[Gao et~al.(2026{\natexlab{c}})Gao, Li, Yuan, Ji, Ma, and
  Wang]{gao_2026_skillreducer}
Yudong Gao, Zongjie Li, Yuanyuan Yuan, Zimo Ji, Pingchuan Ma, and Shuai Wang.
\newblock {SkillReducer}: Optimizing {LLM} agent skills for token efficiency.
\newblock arXiv preprint arXiv:2603.29919, 2026{\natexlab{c}}.

\bibitem[Gautam et~al.(2026)Gautam, Radhakrishna, and
  Gulwani]{gautam_2026_skillaxe}
Srishti Gautam, Arjun Radhakrishna, and Sumit Gulwani.
\newblock {SkillAxe}: Sharpening {LLM}-authored agent skills through
  evaluation-guided self-refinement.
\newblock arXiv preprint arXiv:2606.10546, 2026.

\bibitem[Griffiths and Kalish(2007)]{griffiths_2007_iterated}
Thomas~L. Griffiths and Michael~L. Kalish.
\newblock Language evolution by iterated learning with {Bayesian} agents.
\newblock \emph{Cognitive Science}, 31\penalty0 (3):\penalty0 441--480, 2007.

\bibitem[Huang et~al.(2024)Huang, Chen, Mishra, Zheng, Yu, Song, and
  Zhou]{huang_2024_selfcorrect}
Jie Huang, Xinyun Chen, Swaroop Mishra, Huaixiu~Steven Zheng, Adams~Wei Yu,
  Xinying Song, and Denny Zhou.
\newblock Large language models cannot self-correct reasoning yet.
\newblock In \emph{International Conference on Learning Representations}, 2024.

\bibitem[Huang(2026)]{huang_2026_dsskills}
Wei-Jung Huang.
\newblock Do {LLM}-generated skills make better {AI} data scientists? {A}
  component ablation across data-science workflows.
\newblock arXiv preprint arXiv:2607.07504, 2026.

\bibitem[Jiang et~al.(2026)Jiang, Huang, Xing, Wu, Gao, Cao, Wang, Liu, and
  Li]{jiang_2026_demystifying}
Zhiyuan Jiang, Fangrui Huang, Hanwen Xing, Xander Wu, Yipeng Gao, Rui Cao,
  Mengdi Wang, Shilong Liu, and Yijiang Li.
\newblock Demystifying agent skills: Why they work---until they don't.
\newblock arXiv preprint arXiv:2608.14036, 2026.

\bibitem[Kalish et~al.(2007)Kalish, Griffiths, and
  Lewandowsky]{kalish_2007_iterated}
Michael~L. Kalish, Thomas~L. Griffiths, and Stephan Lewandowsky.
\newblock Iterated learning: Intergenerational knowledge transmission reveals
  inductive biases.
\newblock \emph{Psychonomic Bulletin \& Review}, 14\penalty0 (2):\penalty0
  288--294, 2007.

\bibitem[Khandekar et~al.(2024)Khandekar, Jin, Xiong, Dunn, Applebaum, Anwar,
  Sarfo-Gyamfi, Safranek, Anwar, Zhang, Gilson, Singer, Dave, Taylor, Zhang,
  Chen, and Lu]{khandekar_2024_medcalc}
Nikhil Khandekar, Qiao Jin, Guangzhi Xiong, Soren Dunn, Serina~S Applebaum,
  Zain Anwar, Maame Sarfo-Gyamfi, Conrad~W Safranek, Abid~A Anwar, Andrew
  Zhang, Aidan Gilson, Maxwell~B Singer, Amisha Dave, Andrew Taylor, Aidong
  Zhang, Qingyu Chen, and Zhiyong Lu.
\newblock {MedCalc-Bench}: Evaluating large language models for medical
  calculations.
\newblock In \emph{Advances in Neural Information Processing Systems}, 2024.

\bibitem[Li et~al.(2026{\natexlab{a}})Li, He, Bao, Yang, Chen, Yu, Liang, Su,
  Zhang, and Zhang]{li_2026_skillaligner}
Qinfeng Li, Dalin He, Yuntai Bao, Ying Yang, Ruoxi Chen, Xinyan Yu, Lizhou
  Liang, Ge Su, Wenqi Zhang, and Xuhong Zhang.
\newblock {SkillAligner}: Treating retrieved skills as adaptable drafts at
  execution time.
\newblock arXiv preprint arXiv:2608.06880, 2026{\natexlab{a}}.

\bibitem[Li et~al.(2026{\natexlab{b}})Li, Liu, Zhao, Li, Wang, Shao, Liu, and
  Shou]{li_2026_skillsv}
Tao Li, Junfeng Liu, Qinghua Zhao, Yifan Li, Lei Wang, Bo Shao, Xuejun Liu, and
  Linjun Shou.
\newblock What is a skill worth? {S}tructure-aware {S}hapley valuation of agent
  skills.
\newblock arXiv preprint arXiv:2608.04562, 2026{\natexlab{b}}.

\bibitem[Li et~al.(2026{\natexlab{c}})Li, Liu, Chen, You, Di, He, Zheng, Choe,
  Sun, Wang, Tao, Li, Zhao, Geng, Wu, Zhou, Chen, Xing, Li, Zeng, Wang, Wang,
  Chaim, Jiang, Shen, Kong, Liu, Wang, Liu, Li, Lan, Lin, Ye, He, Li, Zhang,
  Gao, Li, Ma, Jing, Wang, Li, Xue, Lyu, He, Tian, Wu, Wang, Gao, Chen, Liu,
  Cheng, Bao, Tong, Xu, Zhuo, Ye, Qi, Li, Liao, Tan, Shi, Tang, Tankasala,
  Yuan, Qian, Tu, Wang, Sun, Wang, Taylor, Yang, Guan, Dong, Zhang, Dillmann,
  chung Lee, and Song]{li_2026_skillsbench}
Xiangyi Li, Yimin Liu, Wenbo Chen, Bingran You, Zonglin Di, Yifeng He, Shenghan
  Zheng, Kyoung~Whan Choe, Jiankai Sun, Shuyi Wang, Chujun Tao, Binxu Li,
  Xuandong Zhao, Hejia Geng, Xiaojun Wu, Junwei Zhou, Xiaokun Chen, Hanwen
  Xing, Yubo Li, Qunhong Zeng, Di Wang, Yuanli Wang, Roey~Ben Chaim, Penghao
  Jiang, Haotian Shen, Luyang Kong, Xinyi Liu, Runhui Wang, Xuanqing Liu,
  Jiachen Li, Xin Lan, Yueqian Lin, Wengao Ye, Junwei He, Songlin Li, Yue
  Zhang, Yipeng Gao, Yijiang Li, Ze Ma, Liqiang Jing, Tianyu Wang, Kaixin Li,
  Yiqi Xue, Haoran Lyu, Yizhuo He, Yuchen Tian, Shutong Wu, Bowei Wang, Yixuan
  Gao, Bo Chen, Litong Liu, Sikai Cheng, Jiajun Bao, Shuaicheng Tong, Shuwen
  Xu, Terry~Yue Zhuo, Tinghan Ye, Qi Qi, Miao Li, Longtai Liao, Zelin Tan,
  Chang Shi, Xilin Tang, Srinath Tankasala, Boqin Yuan, Yaoyao Qian, Jianhong
  Tu, Chenguang Wang, Yizhou Sun, Wei Wang, Aaron Taylor, Ziyue Yang, Changkun
  Guan, Zhikang Dong, Xinyu Zhang, Steven Dillmann, Han chung Lee, and Dawn
  Song.
\newblock {SkillsBench}: Benchmarking how well agent skills work across diverse
  tasks.
\newblock arXiv preprint arXiv:2602.12670, 2026{\natexlab{c}}.

\bibitem[Liu et~al.(2026)Liu, Zhang, Zhong, Liu, Wang, and
  Wei]{liu_2026_skillshapley}
Chang Liu, Yuqi Zhang, Yiman Zhong, Boyi Liu, Hengjun Wang, and Shuyue Wei.
\newblock {SkillShapley}: Boundary-adaptive {S}hapley valuation for skill step
  attribution in {LLM} agents.
\newblock arXiv preprint arXiv:2608.13173, 2026.

\bibitem[Madaan et~al.(2023)Madaan, Tandon, Gupta, Hallinan, Gao, Wiegreffe,
  Alon, Dziri, Prabhumoye, Yang, Gupta, Majumder, Hermann, Welleck,
  Yazdanbakhsh, and Clark]{madaan_2023_selfrefine}
Aman Madaan, Niket Tandon, Prakhar Gupta, Skyler Hallinan, Luyu Gao, Sarah
  Wiegreffe, Uri Alon, Nouha Dziri, Shrimai Prabhumoye, Yiming Yang, Shashank
  Gupta, Bodhisattwa~Prasad Majumder, Katherine Hermann, Sean Welleck, Amir
  Yazdanbakhsh, and Peter Clark.
\newblock {Self-Refine}: Iterative refinement with self-feedback.
\newblock In \emph{Advances in Neural Information Processing Systems}, 2023.

\bibitem[{Nous Research}(2026)]{nousresearch_2026_hermes}
{Nous Research}.
\newblock Skills system.
\newblock Hermes Agent documentation,
  \url{https://hermes-agent.nousresearch.com/docs/user-guide/features/skills},
  2026.
\newblock Accessed 2026-10-14.

\bibitem[{OpenAI}(2026{\natexlab{a}})]{openai_2026_gpt56}
{OpenAI}.
\newblock {GPT-5.6} system card.
\newblock OpenAI Deployment Safety Hub,
  \url{https://deploymentsafety.openai.com/gpt-5-6}, 2026{\natexlab{a}}.
\newblock Published July 9, 2026.

\bibitem[{OpenAI}(2026{\natexlab{b}})]{openai_2026_gpt61sol}
{OpenAI}.
\newblock Addendum to {GPT-6 Astra} system card: {GPT-6.1 Sol}.
\newblock OpenAI Deployment Safety Hub,
  \url{https://deploymentsafety.openai.com/gpt-6-1-sol}, 2026{\natexlab{b}}.
\newblock Published September 29, 2026.

\bibitem[Ouyang et~al.(2026)Ouyang, Xiao, Gu, and Zhang]{ouyang_2026_skcc}
Yipeng Ouyang, Yi Xiao, Yuhao Gu, and Xianwei Zhang.
\newblock {SkCC}: Portable and secure skill compilation for cross-framework
  {LLM} agents.
\newblock arXiv preprint arXiv:2605.03353, 2026.

\bibitem[Parmar et~al.(2024)Parmar, Patel, Varshney, Nakamura, Luo, Mashetty,
  Mitra, and Baral]{parmar_2024_logicbench}
Mihir Parmar, Nisarg Patel, Neeraj Varshney, Mutsumi Nakamura, Man Luo, Santosh
  Mashetty, Arindam Mitra, and Chitta Baral.
\newblock {LogicBench}: Towards systematic evaluation of logical reasoning
  ability of large language models.
\newblock In \emph{Annual Meeting of the Association for Computational
  Linguistics}, 2024.

\bibitem[Peng et~al.(2026)Peng, Yang, Fu, Zhang, Zhang, Zhao, Su, Wu, Li, Zhu,
  Zhao, and Lei]{peng_2026_sso}
Siran Peng, Cuiyu Yang, Tianyu Fu, Tianshuo Zhang, Haoyuan Zhang, Weisong Zhao,
  Anyang Su, Minghui Wu, Huiying Li, Xiangyu Zhu, Chenxu Zhao, and Zhen Lei.
\newblock Self-supervised skill optimization.
\newblock arXiv preprint arXiv:2607.28777, 2026.

\bibitem[Perez et~al.(2025)Perez, Kova{\v{c}}, L{\'e}ger, Colas, Molinaro,
  Derex, Oudeyer, and Moulin-Frier]{perez_2024_telephone}
J{\'e}r{\'e}my Perez, Grgur Kova{\v{c}}, Corentin L{\'e}ger, C{\'e}dric Colas,
  Gaia Molinaro, Maxime Derex, Pierre-Yves Oudeyer, and Cl{\'e}ment
  Moulin-Frier.
\newblock When {LLMs} play the telephone game: Cultural attractors as
  conceptual tools to evaluate {LLMs} in multi-turn settings.
\newblock In \emph{International Conference on Learning Representations}, 2025.

\bibitem[{Qwen Team}(2026{\natexlab{a}})]{qwen3.8flashnext}
{Qwen Team}.
\newblock {Qwen3.8-Flash-Next}: A new architecture, towards ultimate
  cost-efficiency.
\newblock \url{https://qwen.ai/blog?id=qwen3.8-flash-next}, 2026{\natexlab{a}}.

\bibitem[{Qwen Team}(2026{\natexlab{b}})]{qwen37plus}
{Qwen Team}.
\newblock {Qwen3.7-Plus}: Multimodal agent intelligence.
\newblock \url{https://qwen.ai/blog?id=qwen3.7-plus}, 2026{\natexlab{b}}.

\bibitem[Ren et~al.(2024)Ren, Guo, Qiu, Wang, and
  Sutherland]{ren_2024_biasamplification}
Yi Ren, Shangmin Guo, Linlu Qiu, Bailin Wang, and Danica~J. Sutherland.
\newblock Bias amplification in language model evolution: An iterated learning
  perspective.
\newblock In \emph{Advances in Neural Information Processing Systems}, 2024.

\bibitem[Shen et~al.(2026)Shen, Cheng, Ma, Turcan, Zhang, and
  Ma]{shen_2026_skillfoundry}
Shuaike Shen, Wenduo Cheng, Mingqian Ma, Alistair Turcan, Martin~Jinye Zhang,
  and Jian Ma.
\newblock {SkillFoundry}: Building self-evolving agent skill libraries from
  heterogeneous scientific resources.
\newblock In \emph{Conference on Language Modeling}, 2026.

\bibitem[Shinn et~al.(2023)Shinn, Cassano, Gopinath, Narasimhan, and
  Yao]{shinn_2023_reflexion}
Noah Shinn, Federico Cassano, Ashwin Gopinath, Karthik Narasimhan, and Shunyu
  Yao.
\newblock {Reflexion}: Language agents with verbal reinforcement learning.
\newblock In \emph{Advances in Neural Information Processing Systems}, 2023.

\bibitem[Shumailov et~al.(2024)Shumailov, Shumaylov, Zhao, Papernot, Anderson,
  and Gal]{shumailov_2024_collapse}
Ilia Shumailov, Zakhar Shumaylov, Yiren Zhao, Nicolas Papernot, Ross Anderson,
  and Yarin Gal.
\newblock {AI} models collapse when trained on recursively generated data.
\newblock \emph{Nature}, 631:\penalty0 755--759, 2024.

\bibitem[Su et~al.(2026)Su, Long, Ai, He, Tang, Wang, Tu, Wang, and
  Liu]{su_2026_sra}
Weihang Su, Jianming Long, Qingyao Ai, Qiaozhi He, Yichen Tang, Changyue Wang,
  Yiteng Tu, Yingbo Wang, and Yiqun Liu.
\newblock Skill retrieval augmentation for agentic {AI}.
\newblock arXiv preprint arXiv:2604.24594, 2026.

\bibitem[Wang et~al.(2024)Wang, Xie, Jiang, Mandlekar, Xiao, Zhu, Fan, and
  Anandkumar]{wang_2023_voyager}
Guanzhi Wang, Yuqi Xie, Yunfan Jiang, Ajay Mandlekar, Chaowei Xiao, Yuke Zhu,
  Linxi Fan, and Anima Anandkumar.
\newblock {Voyager}: An open-ended embodied agent with large language models.
\newblock \emph{Transactions on Machine Learning Research}, 2024.

\bibitem[Wang et~al.(2026{\natexlab{a}})Wang, Zhao, Li, Chen, Chen, Peng, and
  He]{wang_2026_skillrevisions}
Jiajie Wang, Yutong Zhao, Tianlin Li, Huashan Chen, Jinfu Chen, Kebin Peng, and
  Sen He.
\newblock Agent skill evolution: How revisions affect coding agents.
\newblock arXiv preprint arXiv:2610.04832, 2026{\natexlab{a}}.

\bibitem[Wang et~al.(2025{\natexlab{a}})Wang, Li, Song, Xu, Tang, Zhuge, Pan,
  Song, Li, Singh, Tran, Li, Ma, Zheng, Qian, Shao, Muennighoff, Zhang, Hui,
  Lin, Brennan, Peng, Ji, and Neubig]{wang_2024_openhands}
Xingyao Wang, Boxuan Li, Yufan Song, Frank~F. Xu, Xiangru Tang, Mingchen Zhuge,
  Jiayi Pan, Yueqi Song, Bowen Li, Jaskirat Singh, Hoang~H. Tran, Fuqiang Li,
  Ren Ma, Mingzhang Zheng, Bill Qian, Yanjun Shao, Niklas Muennighoff, Yizhe
  Zhang, Binyuan Hui, Junyang Lin, Robert Brennan, Hao Peng, Heng Ji, and
  Graham Neubig.
\newblock {OpenHands}: An open platform for {AI} software developers as
  generalist agents.
\newblock In \emph{International Conference on Learning Representations},
  2025{\natexlab{a}}.

\bibitem[Wang et~al.(2026{\natexlab{b}})Wang, Yan, Bi, Yan, Tresp, and
  Ma]{wang_2026_metaskillevolve}
Zefeng Wang, Minxi Yan, Jinhe Bi, Sikuan Yan, Volker Tresp, and Yunpu Ma.
\newblock {MetaSkill-Evolve}: Recursive self-improvement of {LLM} agents via
  two-timescale meta-skill evolution.
\newblock arXiv preprint arXiv:2607.05297, 2026{\natexlab{b}}.

\bibitem[Wang et~al.(2025{\natexlab{b}})Wang, Gandhi, Neubig, and
  Fried]{wang_2025_programmatic}
Zora~Zhiruo Wang, Apurva Gandhi, Graham Neubig, and Daniel Fried.
\newblock Inducing programmatic skills for agentic tasks.
\newblock In \emph{Conference on Language Modeling}, 2025{\natexlab{b}}.

\bibitem[Wang et~al.(2025{\natexlab{c}})Wang, Mao, Fried, and
  Neubig]{wang_2024_workflow}
Zora~Zhiruo Wang, Jiayuan Mao, Daniel Fried, and Graham Neubig.
\newblock Agent workflow memory.
\newblock In \emph{International Conference on Machine Learning},
  2025{\natexlab{c}}.

\bibitem[Wu et~al.(2026)Wu, Gong, Jahangirova, Chen, and
  Zhang]{wu_2026_skilladaptation}
Xinjian Wu, Jingzhi Gong, Gunel Jahangirova, Zhenpeng Chen, and Jie~M. Zhang.
\newblock An empirical study of downstream adaptation for agent skills.
\newblock arXiv preprint arXiv:2607.03238, 2026.

\bibitem[{Xiaomi MiMo Team}(2026)]{mimo2026v25pro}
{Xiaomi MiMo Team}.
\newblock {MiMo-V2.5-Pro}.
\newblock \url{https://huggingface.co/collections/XiaomiMiMo/mimo-v25}, 2026.

\bibitem[Xing et~al.(2026)Xing, Chen, Jin, Wu, Lin, Zhou, Chen, Chen, and
  Xiong]{xing_2026_skillremember}
Qinghua Xing, Yinda Chen, Yaping Jin, Zhenhe Wu, Bohan Lin, Hang Zhou, Xinghao
  Chen, Hanting Chen, and Zhiwei Xiong.
\newblock What should a skill remember? quality--cost trade-offs in cost-aware
  skill rewriting for language model agents.
\newblock arXiv preprint arXiv:2606.09421, 2026.

\bibitem[Yang et~al.(2024)Yang, Wang, Lu, Liu, Le, Zhou, and
  Chen]{yang_2023_opro}
Chengrun Yang, Xuezhi Wang, Yifeng Lu, Hanxiao Liu, Quoc~V Le, Denny Zhou, and
  Xinyun Chen.
\newblock Large language models as optimizers.
\newblock In \emph{International Conference on Learning Representations}, 2024.

\bibitem[Yang et~al.(2026)Yang, Gong, Huang, Yang, Zhou, Huang, Li, Gao, Dai,
  Liu, Qiu, Yang, Chen, Yang, and Luo]{yang_2026_skillopt}
Yifan Yang, Ziyang Gong, Weiquan Huang, Qihao Yang, Ziwei Zhou, Zisu Huang, Yan
  Li, Xuemei Gao, Qi Dai, Bei Liu, Kai Qiu, Yuqing Yang, Dongdong Chen, Xue
  Yang, and Chong Luo.
\newblock {SkillOpt}: Executive strategy for self-evolving agent skills.
\newblock In \emph{Advances in Neural Information Processing Systems}, 2026.

\bibitem[Yu et~al.(2026)Yu, Zhu, Lin, Cui, Ding, and Li]{yu_2026_masa}
Jianxiang Yu, Jiapeng Zhu, Bochen Lin, Qier Cui, Zichen Ding, and Xiang Li.
\newblock Skill is not one-size-fits-all: Model-aware skill alignment for {LLM}
  agents.
\newblock arXiv preprint arXiv:2605.30723, 2026.

\bibitem[Yue et~al.(2025)Yue, Chen, Lu, Zhao, Wang, Yue, Song, and
  Huang]{yue_2025_rlvr}
Yang Yue, Zhiqi Chen, Rui Lu, Andrew Zhao, Zhaokai Wang, Yang Yue, Shiji Song,
  and Gao Huang.
\newblock Does reinforcement learning really incentivize reasoning capacity in
  {LLMs} beyond the base model?
\newblock In \emph{Advances in Neural Information Processing Systems}, 2025.

\bibitem[Zhang et~al.(2025{\natexlab{a}})Zhang, Lazuka, and
  Murag]{anthropic_2025_agentskills}
Barry Zhang, Keith Lazuka, and Mahesh Murag.
\newblock Equipping agents for the real world with {Agent Skills}.
\newblock
  \url{https://www.anthropic.com/research/equipping-agents-for-the-real-world-with-agent-skills},
  2025{\natexlab{a}}.
\newblock Published October 16, 2025.

\bibitem[Zhang et~al.(2026{\natexlab{a}})Zhang, Hu, Lu, Lange, and
  Clune]{zhang_2025_dgm}
Jenny Zhang, Shengran Hu, Cong Lu, Robert~Tjarko Lange, and Jeff Clune.
\newblock Darwin {G}{\"o}del machine: Open-ended evolution of self-improving
  agents.
\newblock In \emph{International Conference on Learning Representations},
  2026{\natexlab{a}}.

\bibitem[Zhang et~al.(2026{\natexlab{b}})Zhang, Hu, Upasani, Ma, Hong,
  Kamanuru, Rainton, Wu, Ji, Li, Thakker, Zou, and Olukotun]{zhang_2025_ace}
Qizheng Zhang, Changran Hu, Shubhangi Upasani, Boyuan Ma, Fenglu Hong,
  Vamsidhar Kamanuru, Jay Rainton, Chen Wu, Mengmeng Ji, Hanchen Li, Urmish
  Thakker, James Zou, and Kunle Olukotun.
\newblock Agentic context engineering: Evolving contexts for self-improving
  language models.
\newblock In \emph{International Conference on Learning Representations},
  2026{\natexlab{b}}.

\bibitem[Zhang et~al.(2026{\natexlab{c}})Zhang, Wang, Cui, Qiu, Li, Zhu, and
  He]{zhang_2026_guardrails}
Xing Zhang, Guanghui Wang, Yanwei Cui, Wei Qiu, Ziyuan Li, Bing Zhu, and
  Peiyang He.
\newblock Guardrails beat guidance: A large-scale study of rules, skills, and
  persistent configuration for coding agents.
\newblock arXiv preprint arXiv:2604.11088, 2026{\natexlab{c}}.

\bibitem[Zhang et~al.(2025{\natexlab{b}})Zhang, Li, Long, Zhang, Lin, Yang,
  Xie, Yang, Liu, Lin, Huang, and Zhou]{zhang_2025_qwen3embedding}
Yanzhao Zhang, Mingxin Li, Dingkun Long, Xin Zhang, Huan Lin, Baosong Yang,
  Pengjun Xie, An Yang, Dayiheng Liu, Junyang Lin, Fei Huang, and Jingren Zhou.
\newblock {Qwen3 Embedding}: Advancing text embedding and reranking through
  foundation models.
\newblock arXiv preprint arXiv:2506.05176, 2025{\natexlab{b}}.

\bibitem[Zhao et~al.(2024)Zhao, Huang, Xu, Lin, Liu, and
  Huang]{zhao_2023_expel}
Andrew Zhao, Daniel Huang, Quentin Xu, Matthieu Lin, Yong-Jin Liu, and Gao
  Huang.
\newblock {ExpeL}: {LLM} agents are experiential learners.
\newblock In \emph{AAAI Conference on Artificial Intelligence}, pages
  19632--19642, 2024.

\bibitem[Zheng et~al.(2025)Zheng, Fatemi, Jin, Wang, Gandhi, Song, Gu,
  Srinivasa, Liu, Neubig, and Su]{zheng_2025_skillweaver}
Boyuan Zheng, Michael~Y. Fatemi, Xiaolong Jin, Zora~Zhiruo Wang, Apurva Gandhi,
  Yueqi Song, Yu Gu, Jayanth Srinivasa, Gaowen Liu, Graham Neubig, and Yu Su.
\newblock {SkillWeaver}: Web agents can self-improve by discovering and honing
  skills.
\newblock arXiv preprint arXiv:2504.07079, 2025.

\bibitem[Zhong et~al.(2026)Zhong, Lu, Ning, Wan, Feng, Ao, Ribeiro, Dreyer,
  Ammirati, and Xiong]{zhong_2026_skilllearnbench}
Shanshan Zhong, Yi Lu, Jingjie Ning, Yibing Wan, Lihan Feng, Yuyi Ao, Leonardo
  F.~R. Ribeiro, Markus Dreyer, Sean Ammirati, and Chenyan Xiong.
\newblock {SkillLearnBench}: Benchmarking continual learning methods for agent
  skill generation on real-world tasks.
\newblock arXiv preprint arXiv:2604.20087, 2026.

\end{thebibliography}
}
\end{document}